\documentclass{article}

\PassOptionsToPackage{numbers, compress}{natbib}

\usepackage[final]{neurips_2024}
\usepackage[utf8]{inputenc} 
\usepackage[T1]{fontenc}    
\usepackage[colorlinks,linkcolor=red,anchorcolor=blue,citecolor=green]{hyperref}
\usepackage{url}            
\usepackage{booktabs}       
\usepackage{amsfonts}       
\usepackage{nicefrac}       
\usepackage{microtype}      
\usepackage{xcolor}         
\usepackage{multirow}
\usepackage{times}
\usepackage{epsfig}
\usepackage{subfloat}
\usepackage{eso-pic}
\usepackage{xspace}
\usepackage{tikz}
\usepackage{bbm}
\usepackage{colortbl,booktabs}
\usepackage{makecell}
\usepackage{diagbox} 
\usepackage{multicol}
\usepackage{array}
\usepackage{makecell}
\usepackage{amsmath, bm}
\usepackage{mathtools}
\usepackage{amsthm}
\usepackage{dsfont}
\usepackage{amssymb}
\usepackage{algorithm}
\usepackage{algpseudocode}
\usepackage{varwidth}
\usepackage{pifont}
\usepackage{makecell}
\usepackage{wrapfig}
\usepackage{enumitem}
\usepackage{caption}
\usepackage{listings}
\usepackage{color}
\usepackage{tikz}
\usetikzlibrary{arrows.meta}
\usepackage{graphicx}

\usepackage{subcaption, multirow, overpic, textpos}

\definecolor{green}{RGB}{0,150,10}
\definecolor{blue}{RGB}{0,148,181}

\title{Training-Free Adaptive Diffusion with Bounded Difference Approximation Strategy}

\author{Hancheng Ye$^{1,*}$, Jiakang Yuan$^{2,*}$, Renqiu Xia$^3$, Xiangchao Yan$^1$, \\ \textbf{Tao Chen}$^2$, \textbf{Junchi Yan}$^3$, \textbf{Botian Shi}$^1$, \textbf{Bo Zhang}$^{1,\ddagger}$ \\
[1mm]
$^1$Shanghai Artificial Intelligence Laboratory\\
$^2$School of Information Science and Technology, Fudan University\\
$^3$School of Artificial Intelligence, Shanghai Jiao Tong University\\[1mm]
\texttt{\small {yehancheng@pjlab.org.cn, jkyuan22@m.fudan.edu.cn, zhangbo@pjlab.org.cn}}
}

\begin{document}

\renewcommand{\thefootnote}{\fnsymbol{footnote}}
\footnotetext{*Equal contribution, $^\ddagger$Corresponding author.}

\maketitle

\begin{abstract}
\vspace{-0.15cm}
  Diffusion models have recently achieved great success in the synthesis of high-quality images and videos. However, the existing denoising techniques in diffusion models are commonly based on step-by-step noise predictions, which suffers from high computation cost, resulting in a prohibitive latency for interactive applications. In this paper, we propose \textit{AdaptiveDiffusion} to relieve this bottleneck by adaptively reducing the noise prediction steps during the denoising process. Our method considers the potential of skipping as many noise prediction steps as possible while keeping the final denoised results identical to the original full-step ones. Specifically, the skipping strategy is guided by the \textit{third}-order latent difference that indicates the stability between timesteps during the denoising process, which benefits the reusing of previous noise prediction results. Extensive experiments on image and video diffusion models demonstrate that our method can significantly speed up the denoising process while generating identical results to the original process, achieving up to an average $2\sim$ $5\times$ speedup without quality degradation. The code is available at \url{https://github.com/UniModal4Reasoning/AdaptiveDiffusion}.
\end{abstract}

\section{Introduction}
\vspace{-0.15cm}

Recently, Diffusion models~\citep{balaji2022ediffi,ho2020denoising,nichol2021improved,rombach2022high} have emerged as a powerful tool for synthesizing high-quality images and videos. Their capability to generate realistic and detailed visual content has made them a popular choice in various applications, ranging from artistic creation to data augmentation, \textit{e.g.}, Midjourney, Sora~\citep{videoworldsimulators2024}, etc. However, the conventional denoising techniques employed in these models involve step-by-step noise predictions, which are computationally intensive and lead to significant latency, \textit{e.g.}, taking tens seconds for SDXL~\citep{podell2023sdxl} to generate a high-quality image of 1024x1024 resolutions. Diffusion acceleration as an effective technique has been deeply explored recently, which mainly focuses on three paradigms: (1) reducing sampling steps~\citep{song2020denoising, kong2021fast, lu2022dpm, salimans2021progressive, lu2022dpm++}, (2) optimizing model architecture~\citep{li2023autodiffusion, tang2023deediff, fang2024structural, ma2023deepcache} and (3) parallelizing inference~\citep{li2024distrifusion,shih2023paradigms}. 

\begin{figure*}[t]
\vspace{-0.85cm}
    \centering
    \resizebox{0.93\linewidth}{!}{
    \includegraphics{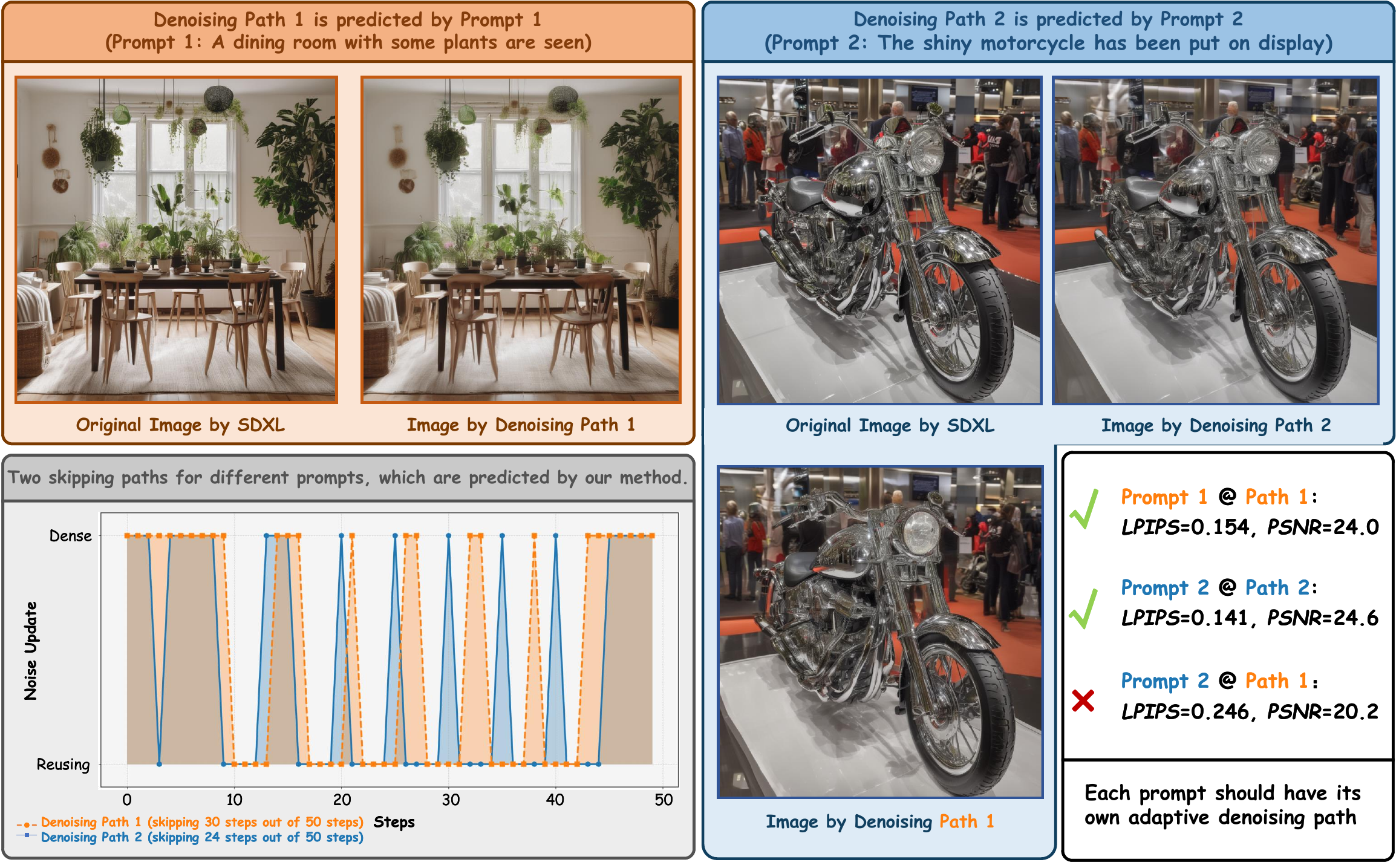}}
    \vspace{-0.15cm}
    \caption{\small Different prompts may have different denoising paths to generate the high-quality image. For Prompt 1, we only need 20 steps out of 50 steps for noise predictions to generate an almost lossless image, while for Prompt 2, we need 26 steps out of 50 steps to achieve an almost lossless image.}
    \label{fig:adaptive_example}
    \vspace{-0.35cm}
\end{figure*}

Currently, most strategies are designed based on a fixed acceleration mode for all prompt data. However, in our experiments, it is observed that different prompts may require different steps of noise prediction to achieve the same content as the original denoising process, as presented in Fig.~\ref{fig:adaptive_example}. Here, we compare the denoising paths using two different prompts for SDXL~\citep{podell2023sdxl}, both of which preserve rich content against the original full-step generation results. The \textit{denoising path} denotes a bool-type sequence, where each element represents whether to infer the noise from the noise prediction model. It can be observed that Prompt 2 needs more steps to generate an almost lossless image than Prompt 1 (A lower LPIPS~\citep{zhang2018perceptual} value means more similarity between two images generated by the original strategy and our strategy). Therefore, it is necessary to explore a \textbf{prompt-adaptive} acceleration paradigm to consider the denoising diversity between different prompts.

Motivated by this observation, in this paper, we deeply dive into the skipping scheme for the noise prediction model and propose AdaptiveDiffusion, a novel approach that adaptively accelerates the generation process according to different input prompts. The fundamental concept behind AdaptiveDiffusion is to adaptively reduce the number of noise prediction steps according to different input prompts during the denoising process, and meanwhile maintain the quality of the final output. The key insight driving our method is that \textbf{the redundancy of noise prediction is highly related to the \textit{third}-order differential distribution between temporally-neighboring latents}. This relation can be leveraged to design an effective skipping strategy, allowing us to decide when to reuse previous noise prediction results and when to proceed with new calculations. Our approach utilizes the \textit{third}-order latent difference to assess the redundancy of noise prediction at each timestep, reflecting our strategy's dependency on input information, thus achieving a prompt-adaptive acceleration paradigm.

Extensive experiments conducted on both image and video diffusion models demonstrate the effectiveness of AdaptiveDiffusion. The results show that our method can achieve up to a 5.6x speedup in the denoising process with better preservation quality. This improvement in acceleration quality opens up new possibilities for the application of diffusion models in real-time and interactive environments.

In summary, AdaptiveDiffusion represents a substantial advancement in adaptively efficient diffusion, offering a practical solution to the challenge of high computational costs associated with sequentially denoising techniques. The main contribution is threefold: (1) To our best knowledge, our method is the first to explore the adaptive diffusion acceleration from the step number reduction of noise predictions that makes different skipping paths for different prompts. (2) We propose a novel approach, namely AdaptiveDiffusion, which develops a plug-and-play criterion to decide whether the noise prediction should be inferred or reused from the previous noise results. (3) Extensive experiments conducted on various diffusion models~\citep{podell2023sdxl, rombach2022high, zhang2023i2vgen, wang2023modelscope} and tasks~\citep{deng2009imagenet, lin2014microsoft, xu2016msr, fan2023aigcbench} demonstrate the superiority of our AdaptiveDiffusion to the existing acceleration methods in the trade-off among efficiency, performance and generalization ability.
\section{Related Works}
\vspace{-0.25cm}
\subsection{Diffusion Models}
\vspace{-0.15cm}

Diffusion models~\citep{balaji2022ediffi,ho2020denoising,nichol2021improved,rombach2022high,fei2023generative} have achieved great success and served as a milestone in content generation. As a pioneer, Denoising Diffusion Probabilistic Models (DDPMs)~\citep{ho2020denoising} generate higher-quality images compared to generative adversarial networks (GANs)~\citep{li2020gan,choi2020starganv2} through an iterative denoising process. To improve the efficiency of DDPM, Latent Diffusion Models (LDMs)~\citep{rombach2022high} perform forward and reverse processes in a latent space of lower dimensionality which further evolves into Stable Diffusion (SD) family~\citep{sauer2024fast,podell2023sdxl}. Recently, video diffusion models~\citep{blattmann2023stable,zhang2023i2vgen,bar2024lumiere,wang2023modelscope} have attracted increasing attention, especially after witnessing the success of Sora~\citep{videoworldsimulators2024}. Stable Video Diffusion (SVD)~\citep{blattmann2023stable} introduces a three-stage training pipeline and obtains a video generation model with strong motion representation. I2VGen-XL~\citep{zhang2023i2vgen} first obtains a model with multi-level feature extraction ability, then enhances the resolution and injects temporal information in the second stage. Despite the high quality achieved by diffusion models, the inherent nature of the reverse process which needs high computational cost slows down the inference process.

\vspace{-0.25cm}
\subsection{Accelerating Diffusion Models}
\vspace{-0.15cm}

Current works accelerate diffusion models can be divided into the following aspects. \textit{(1) Reducing Sampling Steps}~\citep{song2020denoising,lu2022dpm,lu2022dpm++,zhang2022fast,song2023consistency,luo2023latent,lin2024sdxl,sauer2023adversarial,meng2023distillation,moon2023early,lyu2022accelerating}. DDIM~\citep{song2020denoising} optimizes sampling steps by exploring a non-Markovian process. Further studies~\citep{lu2022dpm,lu2022dpm++,zhang2022fast} such as DPM-Solver~\citep{lu2022dpm} propose different solvers for diffusion SDEs and ODEs to reduce sampling steps. Another way to optimize sampling steps is to train few-step diffusion models by distillation~\citep{song2023consistency,luo2023latent,lin2024sdxl}. Among them, consistency models~\citep{song2023consistency,luo2023latent} directly map noise to data to enable one-step generation. Other works~\citep{lin2024sdxl,sauer2023adversarial} explore progressive distillation or adversarial distillation to effectively reduce the reverse steps. Besides, some works~\citep{lyu2022accelerating,tang2023deediff,moon2023early} introduce early stop mechanism into diffusion models. \textit{(2) Optimizing Model Architecture}~\citep{fang2024structural,li2023snapfusion,yang2023diffusion,ma2023deepcache,chen2023sparsevit,ye2024once,liu2024sparse,tang2023torchsparse++}. Another strategy to accelerate diffusion models is to optimize model efficiency to reduce the cost during inference. Diff-pruning~\citep{fang2024structural} compresses diffusion models by employing Taylor expansion over pruned timesteps. DeepCache~\citep{ma2023deepcache} notices the feature redundancy in the denoising process and introduces a cache mechanism to reuse pre-computed features. \textit{(3) Parallel Inference}~\citep{li2024distrifusion,shih2023paradigms,kodaira2023streamdiffusion}. The third line lies in sampling or calculating in a parallel way. ParaDiGMS~\citep{shih2023paradigms} proposes to use Picard iteration to run multiple steps in parallel. DistriFusion~\citep{li2024distrifusion} introduces displaced patch parallelism by reusing the pre-computed feature maps. Compared with the existing paradigms designing a fixed acceleration mode for all input prompts, our method highlights the adaptive acceleration manner with a plug-and-play criterion based on the high-order latent differential distribution, which allows various diffusion models with different sampling schedulers to achieve significant speedup with a negligible performance drop and deployment cost.

\vspace{-0.25cm}
\section{The Proposed Approach}

\begin{figure*}[t]
\vspace{-0.55cm}
    \centering
    \resizebox{0.98\linewidth}{!}{
    \includegraphics{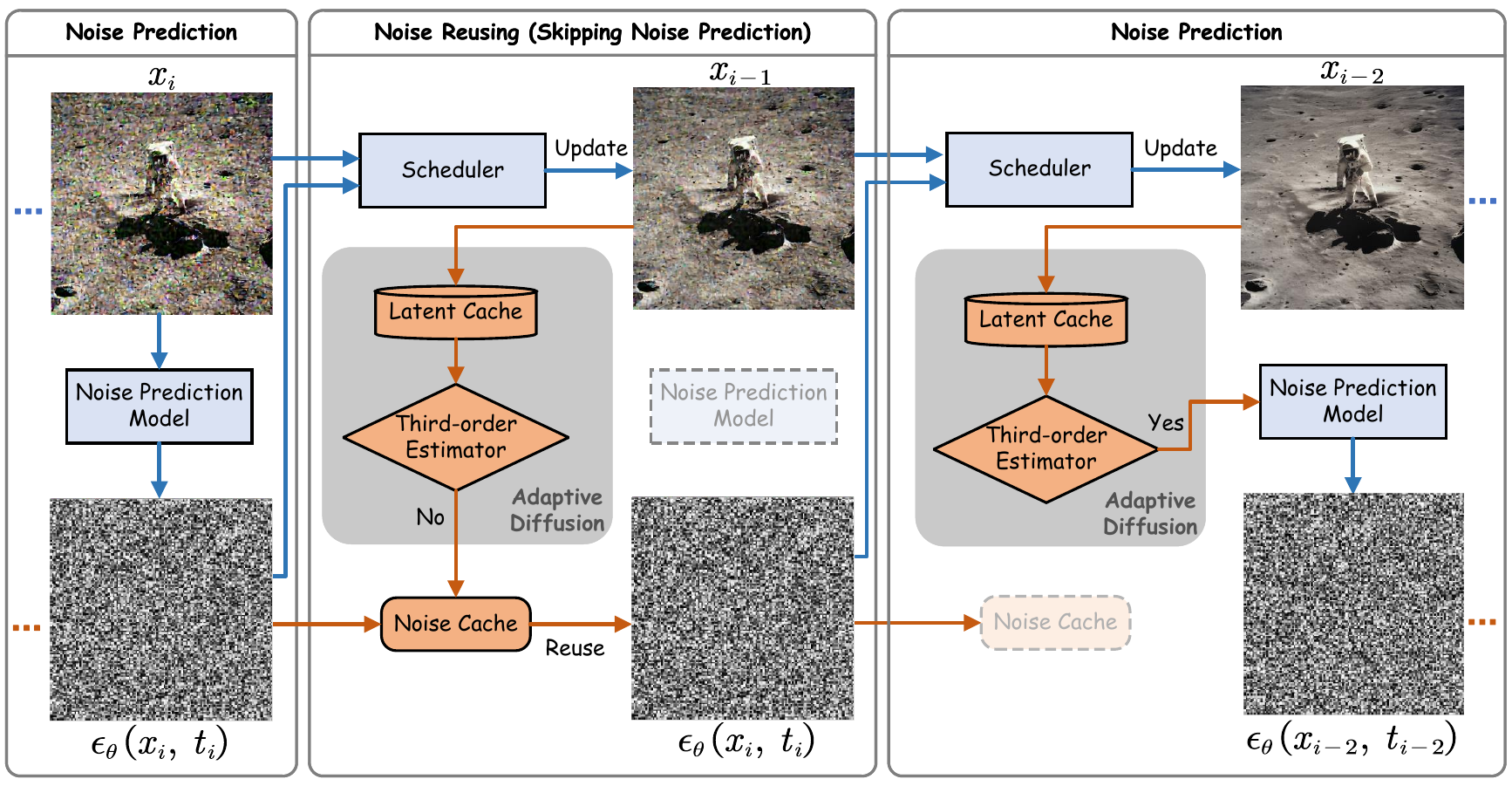}}
    \vspace{-0.25cm}
    \caption{\small Denoising process of the proposed AdaptiveDiffusion: We design a third-order estimator (Refer to Sec.~\ref{sec:third_order} for details), which can find the redundancy between neighboring timesteps, and thus, the noise prediction model can be skipped or inferred according to the indicate from the estimator, achieving the adaptive diffusion process. Note that the timestep and text information embeddings are not shown for the sake of brevity.}
    \label{fig:framwork}
    \vspace{-0.45cm}
\end{figure*}

\begin{figure}[t]
\vspace{-0.55cm}
\centering
\includegraphics[width=0.98\linewidth]{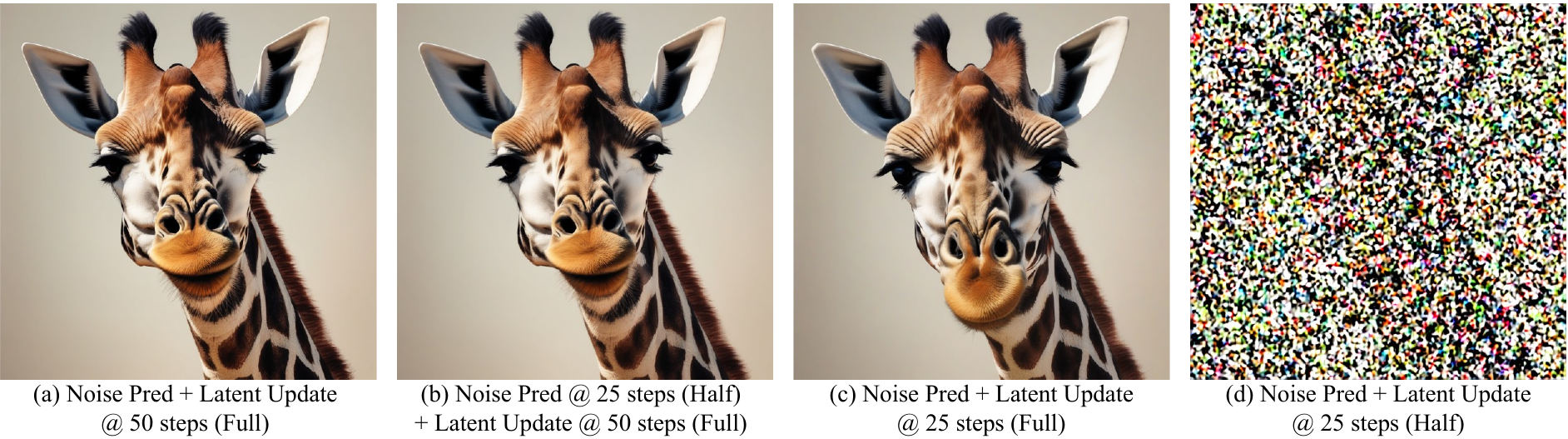}
\vspace{-0.15cm}
\caption{\small Different update strategies. (a) The default SDXL~\citep{podell2023sdxl} samples 50 steps of noise prediction followed by the latent update process. (b) Our AdaptiveDiffusion skips 25 steps of noise prediction according to the \textit{third}-order estimator, while the latent is fully updated at all 50 steps. (c) SDXL samples 25 steps of the noise prediction and latent update process. (d) The default SDXL skips 25 steps of both noise prediction and latent update from its sampled 50 steps.}
\label{fig:update_importance}
\vspace{-0.45cm}
\end{figure}

\vspace{-0.20cm}
\subsection{Preliminary}
\vspace{-0.15cm}

\noindent{\textbf{Reverse Denoising Process.}} Diffusion models \citep{ho2020denoising, song2020denoising} are designed to learn two processes with noise addition (known as the forward process) and noise reduction (known as the reverse process). During the inference stage, only the reverse denoising process is adopted that starts from the Gaussian noise $x_T\sim \mathcal{N}(0, I)$ and iteratively denoises the input sample under the injected condition to get the final clean image(s) $x_0$, where $T$ is the predefined number of denoising steps. Specifically, given an intermediate noisy image $x_i$ at timestep $i$ ($i=1, ..., T$), the noise prediction model $\epsilon_\theta$ (\textit{e.g.}, UNet~\citep{ronneberger2015u}) takes $x_i$, timestep $t_i$ and an additional condition $c$ (\textit{e.g.}, text, image, and motion embeddings, etc) as input to approximate the noise distribution in $x_i$. The update from $x_i$ to $x_{i-1}$ is determined by different samplers (\textit{a.k.a}, schedulers) that can be generally formulated as Eq. (\ref{eq:general_update}):

\setlength{\fboxrule}{1.5pt} 
\vspace{-1em}
\begin{equation}\label{eq:general_update}
\small
x_{i-1}=f(i-1) \cdot x_{i}-g(i-1) \cdot \epsilon_{\theta}(x_{i}, t_{i}),\,\,\,\,\,\,\,i=1,\ldots,T, \, 
\end{equation}
where $f(i)$ and $g(i)$ are step-related coefficients derived by specific samplers~\citep{lu2022dpm, zhang2022gddim}. The computation in the update process mainly involves a few element-wise additions and multiplications. Therefore, the main computation cost in the denoising process stems from the inference of noise prediction model $\epsilon_\theta$ \citep{li2024distrifusion}.

\noindent{\textbf{Step Skipping Strategy.}} As proved by previous works~\citep{ma2023deepcache, tang2023deediff, li2023autodiffusion}, features between consecutive timesteps present certain similarities in distribution, thus there exists a set of redundant computations that can be skipped. Previous works usually skip either the whole update process or the partial computation within the noise prediction model at redundant timesteps. However, as visualized in Fig.~\ref{fig:update_importance}, the latent update process in those redundant timesteps may be important to the lossless image generation. Besides, the calculation redundancy of the noise prediction model within each denoising process is still under-explored. To reduce the computation cost from the noise prediction model, we consider directly reducing the number of noise prediction steps from the original denoising process, which will be proved more effective and efficient to accelerate the generation with almost no quality degradation in Sec.~\ref{sec:third_order}. Given a certain timestep $i$ to skip, our skipping strategy for the update process can be formulated using Eq. (\ref{eq:skip_strategy}):
\begin{equation}\label{eq:skip_strategy}
\small
\begin{aligned}
x_{i}&=f(i) \cdot x_{i+1}-g(i) \cdot \epsilon_{\theta}(x_{i+1}, t_{i+1}),\\
x_{i-1}&=f(i-1) \cdot x_{i}-g(i-1) \cdot \epsilon_{\theta}(x_{i+1}, t_{i+1}).
\end{aligned}
\end{equation}

\vspace{-1em}

\subsection{Error Estimation of the Step Skipping Strategy}
\label{sec:error}
\vspace{-0.15cm}

To validate the effectiveness of our step-skipping strategy, we theoretically analyze the upper bound of the error between the original output and skipped output images. For simplicity, we consider skipping one step of noise prediction, \textit{e.g.}, the $i$-th timestep. Specifically, we have the following theorem about the one-step skipping.

\newtheorem{thm}{\bf Theorem}
\begin{thm}\label{thm1}
Given the skipping timestep $i$ ($i > 0$), the original output $x_{i-1}^{\text{ori}}$ and the skipped output $x_{i-1}$, then the following in-equation holds:
\begin{equation}\label{eq:upper-bound}
\begin{aligned}
\small
\varepsilon_{i-1} &= \|x_{i-1}-x_{i-1}^{ori}\| = \mathcal{O}(t_{i}-t_{i+1})+\mathcal{O}(x_{i}-x_{i+1}).
\end{aligned}
\end{equation}
\end{thm}
The proof can be found in Appendix~\ref{app:proof}. From Eq. (\ref{eq:upper-bound}), it can be observed that the error of the noise-skipped output is upper-bounded by the first-order difference between the previous two outputs. Similarly, the error of the continuously skipped output is upper-bounded by the accumulative differences between multiple previous outputs, which can be found in Appendix~\ref{app:two-step} and \ref{app:k-step}.

Therefore, it can be inferred that as the difference between the previous outputs $x$ is continuously minor, it is possible to predict that the noise prediction at the next timestep can be skipped without damaging the output. This inspires us to utilize the distribution of previous outputs to indicate the skip potential of the next-step noise prediction, as detailed in the following section.

\begin{figure}[t]
\vspace{-0.55cm}
  \centering
  \resizebox{\linewidth}{!}{
  \begin{subfigure}[b]{0.45\textwidth}
    \includegraphics[width=\textwidth]{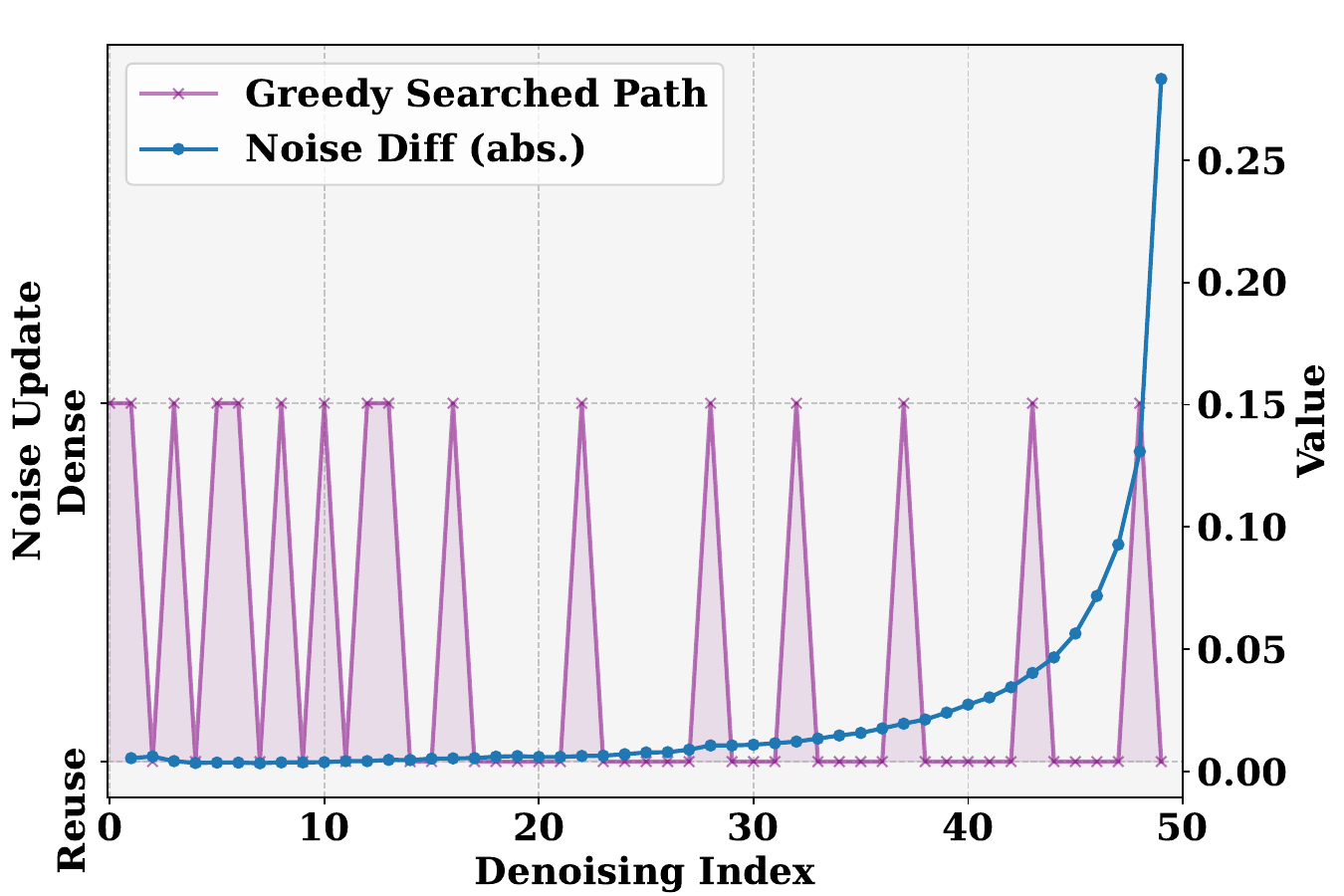}
    \vspace{-1.5em}
    \caption{}
    \label{fig:noise_diff}
  \end{subfigure}
  \begin{subfigure}[b]{0.45\textwidth}
    \includegraphics[width=\textwidth]{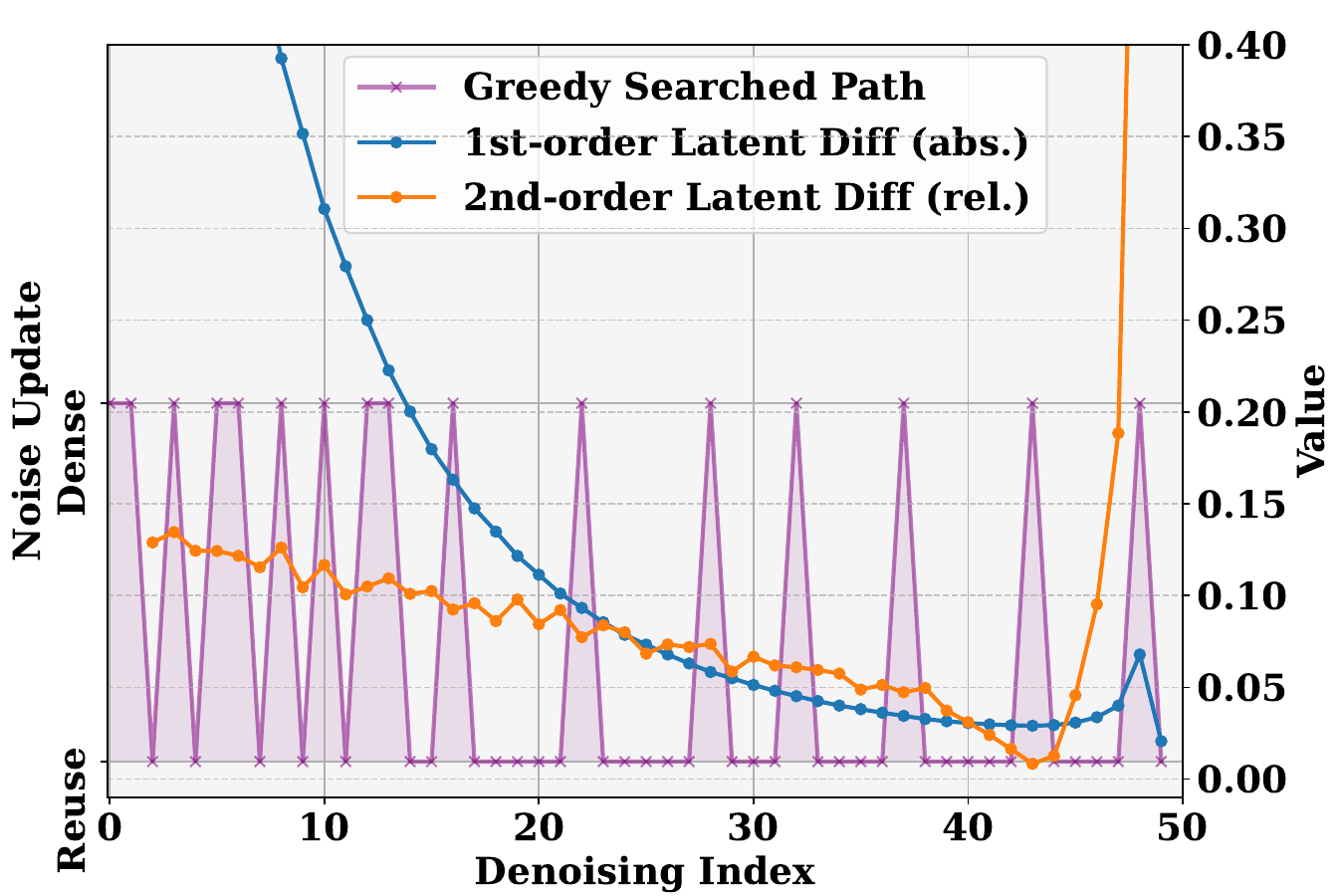}
    \vspace{-1.5em}
    \caption{}
    \label{fig:2nd_diff}
  \end{subfigure}
  \begin{subfigure}[b]{0.45\textwidth}
    \includegraphics[width=\textwidth]{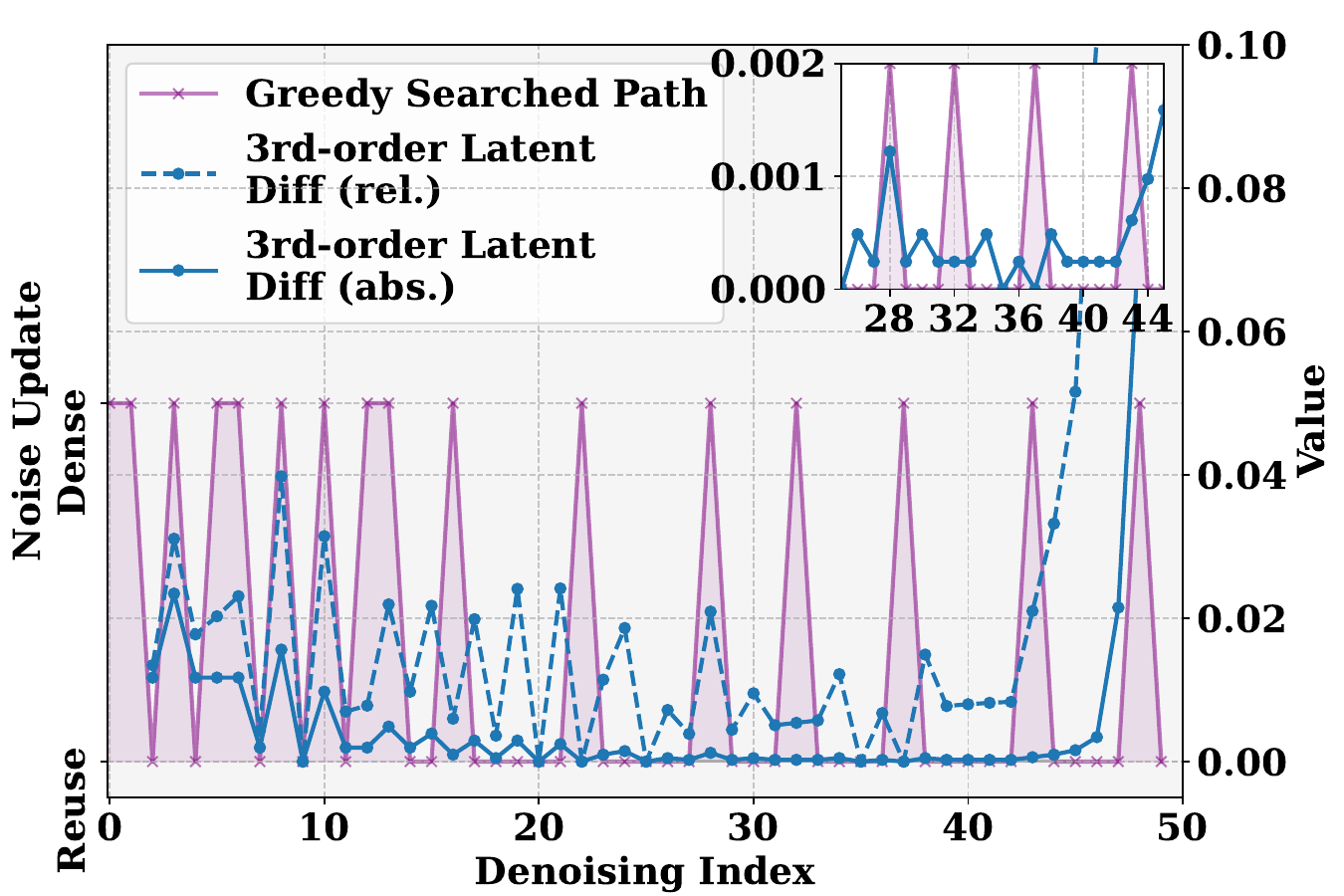}
    \vspace{-1.5em}
    \caption{}
    \label{fig:3rd_diff}
  \end{subfigure}}
  \vspace{-0.65cm}
  \caption{\small The relation between order differential distributions and the searched optimal skipping path for one prompt. (a) The 1st-order noise differential distribution of the original full-step generation shows no relation with the optimal skipping path. (b) The 1st latent differential distribution indicates the distribution of the optimal skipping path but with no explicit mapping with skipping decisions, while the relative 2nd-order latent differential distribution shows a certain skipping signal in its fluctuation, but this signal is buried in the unstable magnitude. (c) The relative 3rd-order latent differential distribution shows a clearer signal for skipping decisions.} \label{fig:estimator_selection}
\vspace{-0.35cm}
\end{figure}

\vspace{-0.20cm}
\subsection{Third-order Estimation Criterion}
\label{sec:third_order}
\vspace{-0.15cm}

\textbf{Observations. }In this section, we take SDXL~\citep{podell2023sdxl} with Euler sampling scheduler as an example to describe the effectiveness of the proposed third-order estimator. Before deriving the third-order estimation criterion, one thing is to calculate the optimal skipping path from the given timestep number, so that we can evaluate the effectiveness of our proposed estimator. Considering the explosive searching cost within a large search space (\textit{e.g.}, searching the optimal path of skipping $N$ steps within $T$ steps for one prompt requires $C_{T}^{N}$ search time cost), we design a greedy search algorithm to approximate the optimal skipping path under different skipping targets, which can be found in Alg. (\ref{alg:greedy}).

Here we randomly take the prompt ``\textit{A bustling 18th-century market scene with vendors, shoppers, and cobblestone streets, all depicted in the detailed oil painting style.}" as an example to visualize the optimal skipping path searched by the greedy search algorithm under the predefined skipping target constraint. Meanwhile, as the skipping error is upper-bounded by the constraint of the first-order latent difference, it inspires us to explore the relationship between the first-order difference distribution and the ideal skipping path.

As shown in Fig. \ref{fig:estimator_selection}, we visualize two types of first-order difference (one in Fig. \ref{fig:noise_diff} representing the noise difference distribution, another in Fig. \ref{fig:2nd_diff} representing latent difference distribution) in the original full-step diffusion to compare with the skipping path. It can be observed that both two distributions of first-order differences are smooth across the denoising process, showing little relationship with the optimal skipping path. A similar insight can be observed in the distribution of the second-order latent difference, as shown in Fig. \ref{fig:2nd_diff}.

However, when considering the third-order latent difference distribution, it presents a significant fluctuation in the original full-step denoising process. As shown in Fig. \ref{fig:3rd_diff}, the distribution of the skipping path is related to the distribution of the third-order latent difference, especially in the early denoising process (around 15 timesteps), where the noise densely updates when the third-order latent difference increases and can be skipped when the third-order difference decreases. As for the later denoising process, when most third-order differences are relatively minor, most noise prediction steps are also skipped. According to the first-order latent difference presented in Fig. \ref{fig:2nd_diff}, the differences between consecutive latent in the early denoising process are significantly larger than those in the later denoising process. Therefore, the precise importance estimation of noise predictions in the early process is much more important and the third-order difference distribution can intuitively serve as the indicator of the noise prediction strategy. The theoretical relation between the third-order derivative of latents and the optimal skipping scheme is analyzed in Appendix \ref{app:theoretical_third}.

\textbf{Criterion. }Based on the above empirical observation of the relationship between the optimal skipping path and the high-order difference distributions, the third-order estimator is proposed to indicate the potential of skipping the noise computation. Specifically, the criterion is formulated as follows:
\begin{equation}\label{eq:criterion}
\small 
\xi \left( x_{i-1} \right) =\left\| \varDelta ^{(3)}x_{i-1} \right\| \ge \delta\|\varDelta x_{i}\|, 
\end{equation}
where $\xi\left( x_{i-1} \right)$ is the indicator that takes $x_{i-1}$ and previous latents $x_{i}, x_{i+1}, x_{i+2}$ as input to estimate whether the next noise prediction can be skipped. If $\xi(x_{i-1})$ returns False, then the noise from the previous step will be reused to update $x_{i-1}$. $\varDelta ^{(3)}x_{i-1}$ denotes the third-order latent difference at timestep $i-1$, \textit{i.e.}, $\varDelta ^{(3)}x_{i-1} = \varDelta ^{(2)}x_{i-1} - \varDelta ^{(2)}x_{i}=\varDelta x_{i-1}-2\varDelta x_{i}+\varDelta x_{i+1}$, and $\varDelta x_i$ is defined as the difference between $x_i$ and $x_{i+1}$ ($i=0, ..., T-1$). $\delta$ is a hyperparameter thresholding the relative scale of $\varDelta ^{(3)} x_{i-1}$. The reason for selecting $\varDelta x_{i}$ is that $\varDelta ^{(3)} x_{i-1}$ actually describes the distance between $(\varDelta x_{i-1} + \varDelta x_{i+1}) / 2$ and $\varDelta x_{i}$. Therefore, it is natural to utilize the relative distance against $\varDelta x_{i}$ to indicate the stability of the denoising process. Fig. \ref{fig:3rd_diff} present a strong relation between $\|\varDelta ^{(3)} x_{i-1} / \varDelta x_{i}\|$ (the blue dashed line) and the optimal skipping path.

\begin{figure}[t]
\vspace{-0.55cm}
\small
  \centering
  \resizebox{\linewidth}{!}{
  \begin{subfigure}[b]{0.4\textwidth}
    \includegraphics[width=\textwidth]{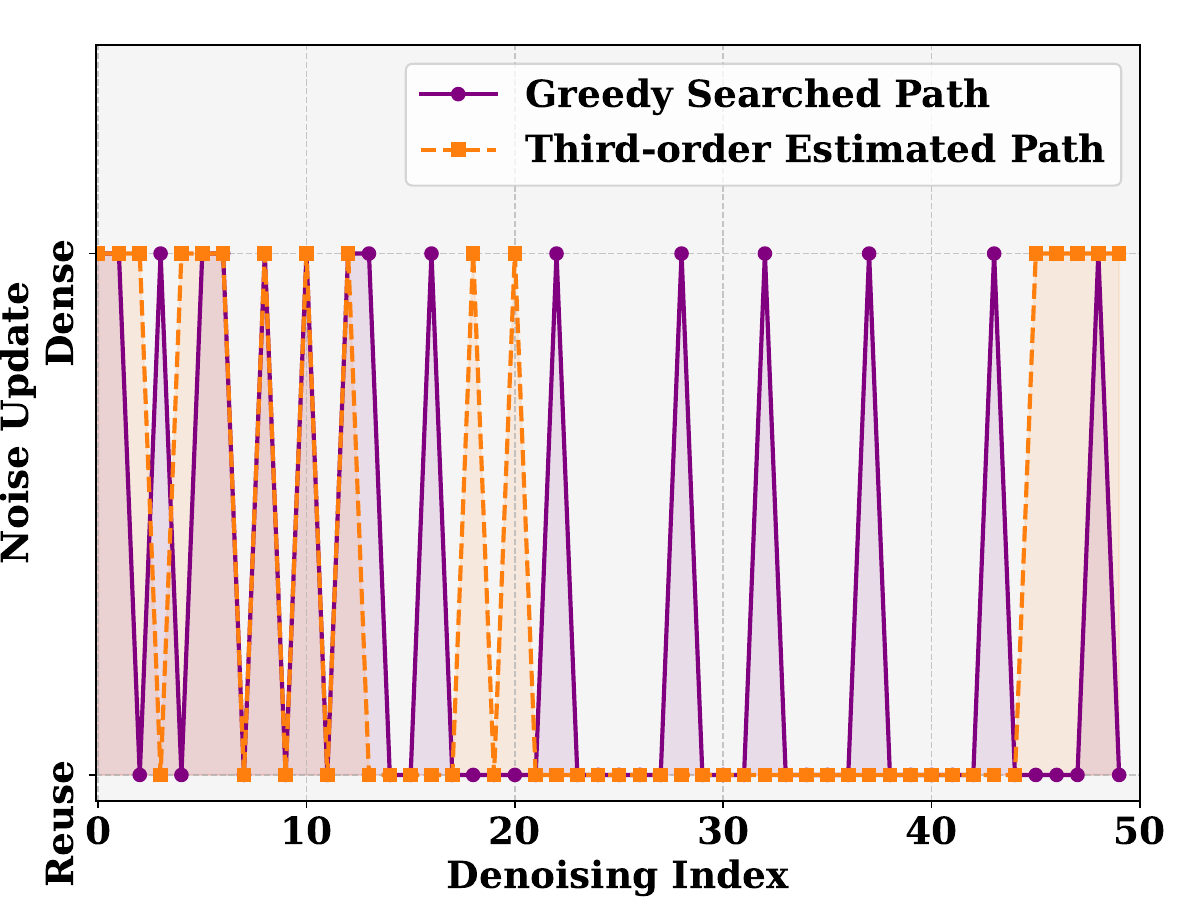}
    \vspace{-1.5em}
    \caption{}
    \label{fig:path_com}
  \end{subfigure}
  \begin{subfigure}[b]{0.45\textwidth}
    \includegraphics[width=\textwidth]{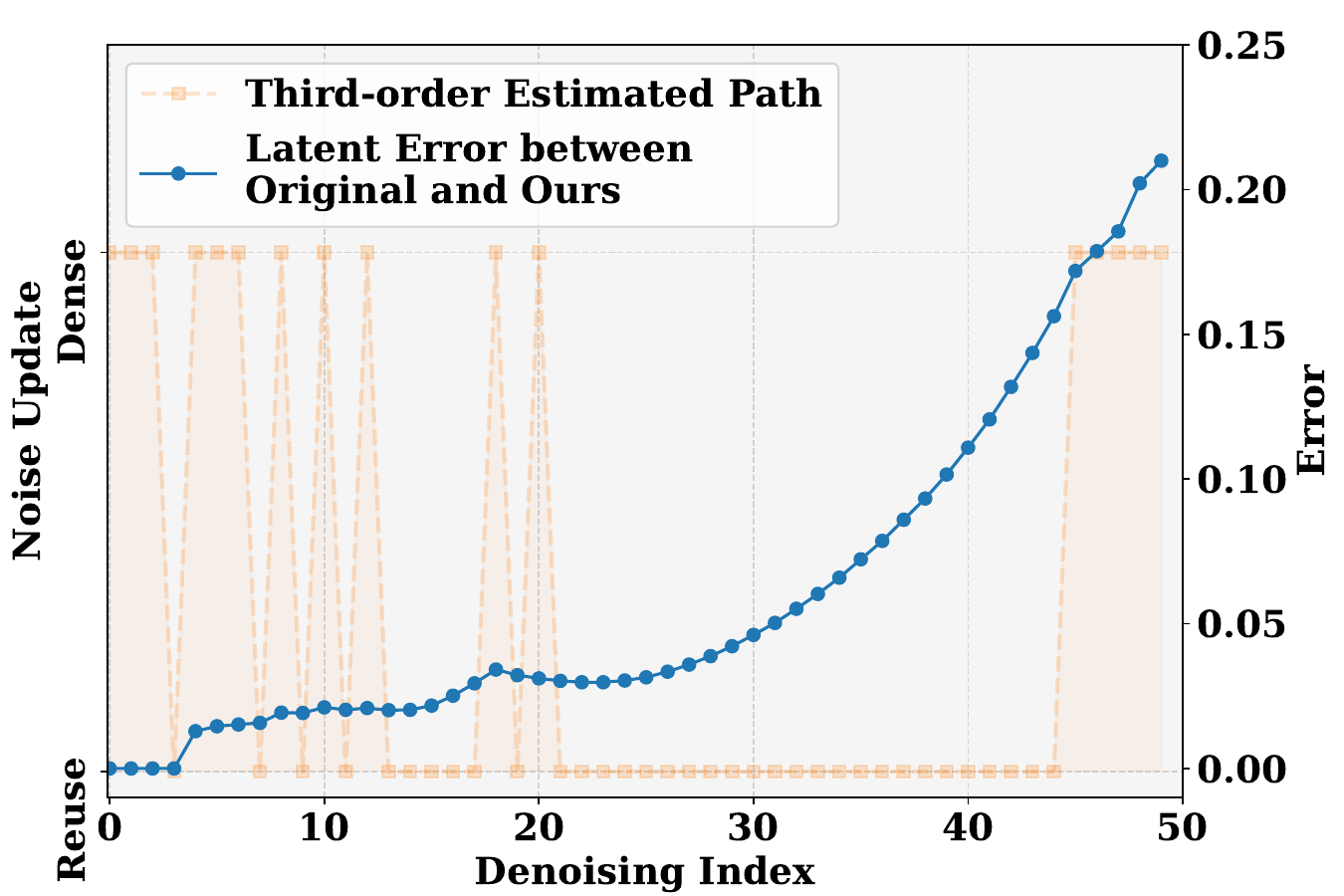}
    \vspace{-1.5em}
    \caption{}
    \label{fig:accum_error}
  \end{subfigure}
  \begin{subfigure}[b]{0.45\textwidth}
    \includegraphics[width=\textwidth]{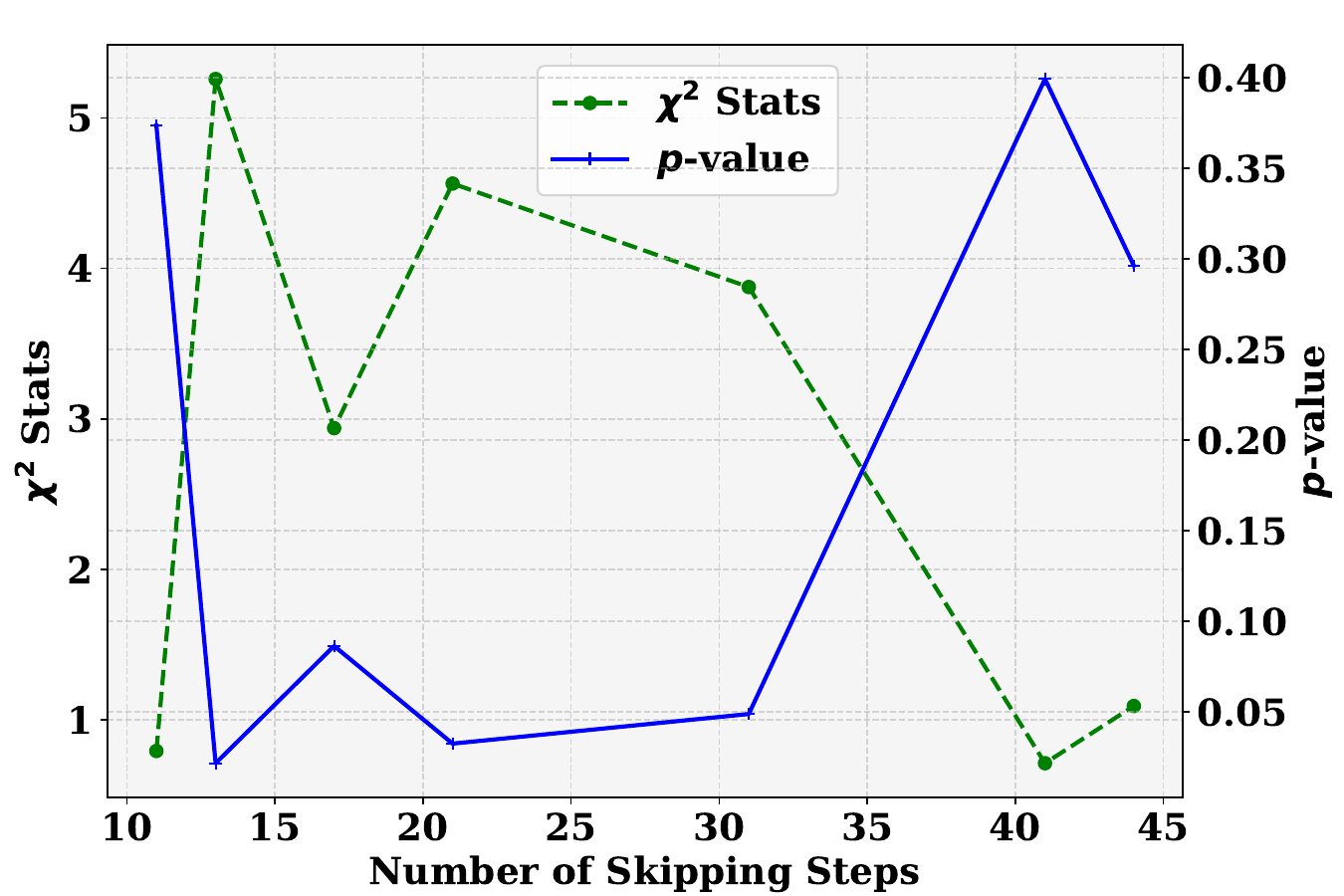}
    \vspace{-1.5em}
    \caption{}
    \label{fig:path_statis}
  \end{subfigure}}
  \vspace{-0.65cm}
  \caption{\small The effectiveness of the proposed \textit{third}-order estimator. (a) The third-order estimated skipping path shares a similar distribution with the optimal skipping path. (b) The latent error between the full-step update path and the estimated skipping path. (c) The $\chi^2$ stats and $p$-value between the greedy searched paths and the third-order estimated paths at different skipping targets.}
  \label{fig:estimator_effectiveness}
\vspace{-0.45cm}
\end{figure}

\textbf{Effectiveness of the Third-order Estimator.} To validate the effectiveness of the proposed third-order estimator, we compare our third-order estimated path with the optimal skipping path searched by the greedy algorithm, which is shown in Fig. \ref{fig:path_com}. It can be observed that the distribution of our estimated path is largely similar to the optimal skipping path. The reason for continuous skipping in the later denoising process is that the third-order difference keeps approaching zero as illustrated in Fig. \ref{fig:3rd_diff}. The accumulative error caused by skipping noise predictions is described in Fig. \ref{fig:accum_error}, where it is observed that the error starts increasing quickly after continuously skipping the noise predictions. Thus, it is vital to introduce another hyperparameter, \textit{i.e.}, the maximum step number of continuous skipping $C_{\text{max}}$, to control the accumulative error. Hyperparameter analyses are described in Sec.~\ref{sec:insightful_analysis}.

Furthermore, we analyze the statistical correlation between the estimated path and the optimal path to test whether the designed criterion is significantly correlated to the optimal skipping criterion. As shown in Fig. \ref{fig:path_statis}, we compute the $\chi^2$ stats and $p$-values under different step numbers of skipping. The results indicate that when the skipping steps are moderate, the estimated skipping path and the optimal skipping path are significantly correlated. For those targeting at small and large numbers of skipping steps, the correlation is statistically insignificant. The test details can be found in Appendix \ref{app:exp_details}. The overall skipping algorithm is shown in Alg. (\ref{alg:adaptive}) of Appendix \ref{app:algorithms}.

\vspace{-0.15cm}
\section{Experiments}
\vspace{-0.15cm}

\vspace{-0.25cm}
\subsection{Experimental Setup}
\label{sec:experimental_setup}
\vspace{-0.15cm}

\textbf{Models.} We conduct experiments in three prompt-based settings including text-to-image (T2I), image-to-video (I2V), and text-to-video (T2V) generation tasks. In addition, we also test the effectiveness of AdaptiveDiffusion on the conditional image generation task. For the T2I task, we use Stable Diffusion-v1-5 (SD-1-5)~\citep{rombach2022high} and Stable Diffusion XL (SDXL)~\citep{podell2023sdxl} and evaluate on three different sampling schedulers (\textit{i.e.}, DDIM~\citep{song2020denoising}, DPM-Solver++~\citep{lu2022dpm++}, and Euler). For the I2V and T2V tasks, we utilize I2VGen-XL~\citep{zhang2023i2vgen} and ModelScopeT2V~\citep{wang2023modelscope} respectively. Note that we use ZeroScope-v2 instead of the original ModelScopeT2V model to generate watermark-free videos. For conditional image generation, we use LDM-4~\citep{rombach2022high} as the baseline model. 

\textbf{Benchmark Datasets.} Following~\citep{ma2023deepcache}, we use ImageNet~\citep{deng2009imagenet} and MS-COCO 2017~\citep{lin2014microsoft} to evaluate the results on class-conditional image generation and T2I tasks, respectively. For the I2V task, we randomly sample 100 prompts and reference images in AIGCBench~\citep{fan2023aigcbench}. For the T2V task, we conduct experiments on a widely-used benchmark MSR-VTT~\citep{xu2016msr} and sample one caption for each video in the validation set as the test prompt. More details can be found in Appendix \ref{app:exp_details}.

\textbf{Comparison Baselines.} We compare AdaptiveDiffusion against DeepCache and Adaptive DPM-Solver in both generation quality and efficiency. Deepcache~\citep{ma2023deepcache} caches high-level features of UNet to update the low-level features at each denoising step, thus reducing the computational cost of UNet. The latter~\citep{lu2022dpm} dynamically adjusts the step size by combining different orders of DPM-Solver. 

\textbf{Evaluation Metrics.} For all tasks, we evaluate our proposed method in both quality and efficiency. We report MACs, latency, and speedup ratio to verify the efficiency. For the image generation task, following previous works~\citep{li2020gan,li2024distrifusion,li2022efficient}, we evaluate image quality with commonly-used metrics, \textit{i.e.}, Peak Signal Noise Ratio (PSNR), Learned Perceptual Image Patch Similarity (LPIPS), and Fréchet Inception Distance (FID). For the video generation task, we use per-frame PSNR and LPIPS to measure the quality of generated videos. Besides, Fréchet Video Distance (FVD)~\citep{unterthiner2019fvd} is also used to quantify the temporal coherence and quality of each frame. \textit{Note that since our method achieves adaptive acceleration results, all reported metrics of our method are averaged across all prompts.}

\textbf{Implementation Details.} We conduct all experiments on RTX 3090 GPUs. For SD-1-5 and SDXL models, the original sampling timesteps $T$ are set as 50, and two hyperparameters are set as: $\delta=0.01$, $C_{\text{max}}=4$. For LDM-4, $T=250, \delta=0.005, C_{\text{max}}=10$. For I2VGen-XL and ModelScopeT2V, $T=50, \delta=0.007, C_{\text{max}}=4$. More details of other methods are listed in Appendix \ref{app:exp_details}.

\vspace{-0.15cm}
\subsection{Main Results}
\label{sec:main_results}
\vspace{-0.15cm}

\subsubsection{Results on Image Generation}
\vspace{-0.15cm}

We first evaluate our method on T2I generation. As shown in Tab.~\ref{tab:image_generation_1}, compared to DeepCache and Adaptive DPM methods, AdaptiveDiffusion achieves both higher quality and efficiency in various settings. For example, AdaptiveDiffusion achieves 0.092 LPIPS on SD-v1-5~\citep{rombach2022high}, generating almost lossless images compared to those generated by the full-step denoising process. Meanwhile, the averaged speedup ratio across all testing prompts achieves 2.01$\times$ when using SDXL and Euler sampling scheduler. In addition, when comparing the generation performance between different models (\textit{e.g.}, SD-1-5 and SDXL~\citep{podell2023sdxl})) and schedulers (\textit{e.g.}, DDIM, DPM-Solver++, and Euler), AdaptiveDiffusion shows stronger generalization capability to adapt to different settings.
\begin{table*}[t]
\vspace{-0.55cm}
    \caption{Quantitative results on MS-COCO 2017~\citep{lin2014microsoft}.}
    \vspace{-0.15cm}
    \label{tab:image_generation_1}
    \centering
    \resizebox{1.0\linewidth}{!}
    {
        \begin{tabular}{llcccccccc}
        \toprule[1.5pt]
        Model  & Scheduler & Method & PSNR $\uparrow$& LPIPS $\downarrow$& FID $\downarrow$ &MACs (T)&Mem (MB)& Latency (s) & Speedup Ratio \\
        \cmidrule{1-10}

        \multirow{6}{*}{SD-1-5}  & \multirow{2}{*}{DDIM}  & Deepcache   &  19.05          &    0.199       & 6.96          & 26  &  3940   &   1.7   & 1.58$\times$ \\
                                  &                        & Ours        &   \textbf{21.74}&  \textbf{0.131}&  \textbf{5.22}&  \textbf{22}   & \textbf{3850}  & \textbf{1.5}     & \textbf{1.77$\times$} \\
        \cmidrule(lr){2-10}
                                 & \multirow{2}{*}{DPM++} &  Adaptive DPM &                 19.0 & 0.195               &     6.38          &   31  & 3896  &   3.5   & 1.25$\times$\\
                                 &                        & Ours         &   \textbf{23.2}              &   \textbf{0.092}             & \textbf{4.06}              &  \textbf{25}   & \textbf{3852}  &   \textbf{2.8}   & \textbf{1.57$\times$}\\
        \cmidrule(lr){2-10}
                                 & \multirow{2}{*}{Euler} & Deepcache              &   18.89         &   0.240        &  7.51         &  26   & 3894  &   1.6    & 1.57$\times$ \\
                                 &                       & Ours                   &   \textbf{20.60}&  \textbf{0.157}& \textbf{6.02} &  \textbf{21}   & \textbf{3848}  &  \textbf{1.3}    & \textbf{1.98$\times$} \\
        \cmidrule(lr){1-10}
                                 
        \multirow{6}{*}{SDXL}   & \multirow{2}{*}{DDIM}  &  Deepcache              & 21.9            &  0.221         & 7.34          &   \textbf{162}  & 14848  &   \textbf{7.4}   & \textbf{1.75$\times$} \\
                                 &                        & Ours                   &\textbf{25.4}    & \textbf{0.141} & \textbf{5.13} &  186   & \textbf{14460}  & 7.8     & 1.66$\times$ \\
        \cmidrule(lr){2-10}
                                 & \multirow{2}{*}{DPM++} & Deepcache              &  21.3           &  0.255         & 8.48          &  \textbf{162}   & 14800  &    \textbf{7.6}  & \textbf{1.74$\times$} \\
                                 &                        & Ours                   &   \textbf{26.1} &  \textbf{0.125}& \textbf{4.59} &  190   & \textbf{14454}  &    8.0  &1.65$\times$\\
        \cmidrule(lr){2-10}
                                 & \multirow{2}{*}{Euler} & Deepcache              &  22.0           &  0.223         &  7.36         &   \textbf{162}  & 14796  &   \textbf{6.3}   & \textbf{2.16$\times$} \\
                                 &                        & Ours                   &   \textbf{24.33}& \textbf{0.168} & \textbf{6.11} &  174   & \textbf{14458}  &6.7      & 2.01$\times$\\
        \bottomrule[1.5pt]
    \end{tabular}
   }
\vspace{-0.45cm}
\end{table*}
\begin{table*}[h]
\vspace{-0.15cm}
    \caption{Quantitative results on ImageNet 256$\times$256~\citep{deng2009imagenet}.}
    \vspace{-0.15cm}
    \label{tab:image_generation}
    \centering
    \resizebox{0.8\linewidth}{!}
    {
        \begin{tabular}{lcccccccc}
        \toprule[1.5pt]
        Model  & Method & PSNR $\uparrow$& LPIPS $\downarrow$& FID $\downarrow$ & Mem (MB) & MACs (T)& Latency (s) & Speedup Ratio \\
        \midrule   
        \multirow{2}{*}{LDM-4}   & Deepcache             &       24.98           &       0.098         &   5.22     & 3030      &  \textbf{5.6}    &   \textbf{1.4}   & \textbf{6.35$\times$} \\
                               & Ours                  &            \textbf{25.79}      &     \textbf{0.090}           &   \textbf{4.91}     &   \textbf{2912}    &  6.5    &   1.6   & 5.56$\times$\\
        \bottomrule[1.5pt]
    \end{tabular}
   }
\vspace{-0.15cm}
\end{table*}

To further verify the effectiveness of our method on different image generation tasks, we also conduct experiments on LDM-4 for the conditional image generation task. As shown in Tab.~\ref{tab:image_generation}, due to the larger timestep setting in the original denoising process, the slow change of latent change allows us to achieve faster generation with almost lossless quality at nearly 5.6x speedup. Compared with Deepcache, the AdpativeDiffusion obtains a better trade-off between generation quality and efficiency.

\vspace{-0.15cm}
\subsubsection{Results on Video Generation}
\vspace{-0.15cm}

We further conduct experiments on more tough tasks, \textit{e.g.}, video generation tasks including I2V and T2V generation. As shown in Tab.~\ref{tab:video}, AdaptiveDiffusion can generate videos of lossless frames and similar video quality with a significant speedup against the original models' full-step generated videos. Specifically, our proposed method can achieve a significantly higher quality in single frame evaluation which can be seen by LPIPS and PSNR (\textit{e.g.}, +6.38dB PSNR compared to DeepCache using I2VGen-XL). On the other hand, AdaptiveDiffusion can ensure temporal consistency as the original models due to the property of lossless acceleration which can be reflected on FVD.

\begin{table}[t]
    \caption{Quantitative results on video generation tasks.}
    \label{tab:video}
    \centering
    \resizebox{1.0\linewidth}{!}
    {
        \begin{tabular}{llccccccc}
        \toprule[1.5pt]
        Model                     & Method & PSNR $\uparrow$ & LPIPS $\downarrow$ & FVD $\downarrow$ & Mem (GB)& MACs (T) & Latency (s) & Speedup Ratio \\
        \midrule
        \multirow{2}{*}{I2VGen-XL}   
        & Deepcache  & 20.82 & 0.212 & 916.0 & 28.0 (+58.4\%) & \textbf{1878} &  \textbf{77} & \textbf{2.00$\times$}  \\
        & Ours       & \textbf{27.20} & \textbf{0.088} & \textbf{274.5} &  \textbf{17.7 (+0.3\textperthousand)} &    1983      &       80        &      1.93$\times$            \\
        \midrule
        \multirow{2}{*}{ModelScopeT2V} 
        & Deepcache & 18.24 & 0.343 & 3383.0 & 14.2 (+50.8\%)  &  1034   &  45 & 1.37$\times$ \\
        & Ours      & \textbf{27.70} & \textbf{0.081} & \textbf{470.6}  & \textbf{8.7 (-7.8\%)} & \textbf{948} &  \textbf{42}   & \textbf{ 1.46$\times$} \\
        \bottomrule[1.5pt]
        \end{tabular}
    }
    
    \vspace{-0.1cm}
\end{table}

\vspace{-0.15cm}
\subsection{Ablation Studies}
\label{sec:insightful_analysis}
\vspace{-0.15cm}

\paragraph{Ablation Study on Skipping Threshold.} As shown in the upper part of Tab.~\ref{tab:ablation}, we first analyze the effect of the skipping threshold $\delta$. It can be observed that when the skipping threshold is relatively small, there will be fewer skipping steps, resulting in a relatively small yet still clear acceleration with a much higher preservation quality. With the threshold gradually increasing, the speedup ratio will be largely improved but at the cost of quality degradation. It can be seen that the image quality does not significantly change with large thresholds (\textit{i.e.}, 0.015 and 0.02). This is because the pre-defined maximum skipping steps prevent the further increase of skip steps. In this paper, we set the skipping threshold to 0.01 which is a better trade-off between the generation quality and inference speed.

\begin{table}[t]
\caption{Ablation studies on hyperparameters using SDXL~\citep{podell2023sdxl}. We conduct the ablation studies using 50-step Euler sampling scheduler for SDXL on COCO2017.}
    \label{tab:ablation}
    \centering
    \resizebox{0.82\linewidth}{!}
    {
        \begin{tabular}{lccccccc}
        \toprule[1.5pt]
        Hyper-params                                  & Value & PSNR $\uparrow$ & LPIPS $\downarrow$ & FID $\downarrow$ & MACs (T) & Latency (s) & Speedup Ratio \\
        \cmidrule(lr){1-8}
        \multirow{5}{*}{\makecell{$\delta$ \\ ($C_{max}$ = 4)}}    
        & 0.005 &   34.3   &  0.023     & 0.96    &   262   &   9.7      &      1.38$\times$         \\                                                   
        & 0.008 &   28.6   & 0.079      & 3.22    &   217   &   7.7      & 1.74$\times$                 \\
        & 0.01  &  24.3 & 0.168  & 6.11   &  174      &      6.7  &       2.01$\times$        \\
        & 0.015 &   21.0  &  0.282     & 9.49    &   119   &   5.1      & 2.64$\times$                 \\
        & 0.02  &  21.0    & 0.289      & 9.79    &   106   &      4.4   &       3.06$\times$       \\
        \cmidrule(lr){1-8}
        \multirow{4}{*}{\makecell{$C_{max}$ \\ ($\delta$ = 0.01)}} 
        & 4     & 24.3 & 0.168  & 6.11 &  174     &   6.7      &   2.01$\times$            \\
        & 6     &  23.3    & 0.217      & 7.74    &  147    &    5.7     & 2.37$\times$              \\
        & 8     & 22.7  & 0.256   & 9.43    &   135   &       5.2  & 2.59$\times$              \\
        & 10    &  22.1    & 0.307 & 11.91     &   123   &     4.7    & 2.86$\times$    \\      \bottomrule[1.5pt]
        \end{tabular}
    }
\vspace{-0.25cm}
\end{table}

\paragraph{Ablation Study on Maximum Skipping Steps.} Further, we conduct ablation studies on the maximum skipping steps $C_{\text{max}}$, as shown in the lower part of Tab.~\ref{tab:ablation}. With the increase of the maximum skipping steps, the quality of generated images continuously decreases. The reason is that when the timestep $t_i$ approaches 0, a large number of denoising steps will be skipped due to the minor value of the third-order latent difference when the max-skip-step is relatively large. The phenomenon reveals that $C_{\text{max}}$ can effectively prevent the continuous accumulation of generated image errors and ensure image quality.

\paragraph{Analysis on Sampling Steps.} To evaluate the effectiveness of AdaptiveDiffusion on few-step sampling. It can be seen from Tab.~\ref{tab:steps} that AdaptiveDiffusion can further accelerate the denoising process under the few-step settings. Note that the hyperparameters (\textit{i.e.}, $\delta$, and $C_{\text{max}}$) should be slightly adjusted according to the original sampling steps due to the varying updating magnitudes in different sampling steps. Specifically, a higher threshold $\delta$ and lower max-skip-step $C_{\text{max}}$ can make better generation quality when reducing the sampling steps.

\begin{table}[h]
\vspace{-0.35cm}
    \caption{Study on few-step sampling. Acceleration results with different original sampling steps using SDXL~\citep{podell2023sdxl} on COCO2017.}
    \label{tab:steps}
    \centering
    \resizebox{0.75\linewidth}{!}
    {
        \begin{tabular}{llccccc}
        \toprule[1.5pt]
        Steps & PSNR $\uparrow$  & LPIPS $\downarrow$ & FID $\downarrow$ & MACs (T) & Latency (s) & Speedup Ratio \\
        \cmidrule(lr){1-7}
        50 steps &      24.3 & 0.168  & 6.11    &    174                      &             6.7                 & 2.01$\times$                                  \\
        25 steps & 32.9  & 0.047 & 1.62 &       128                   &           5.8                 &      1.31$\times$                             \\
        15 steps &  19.9  & 0.122 & 5.06                         &          70                &                        3.1     & 1.38$\times$ \\ 
        10 steps & 29.4 & 0.169 & 8.28 & 53 & 2.5 & 1.21$\times$\\
        \bottomrule[1.5pt]     
        \end{tabular}
    }
\end{table}

\subsection{Visualization Results}
\vspace{-0.15cm}

\begin{figure}[tb!]
\vspace{-0.45cm}
\centering
\includegraphics[width=1.0\linewidth]{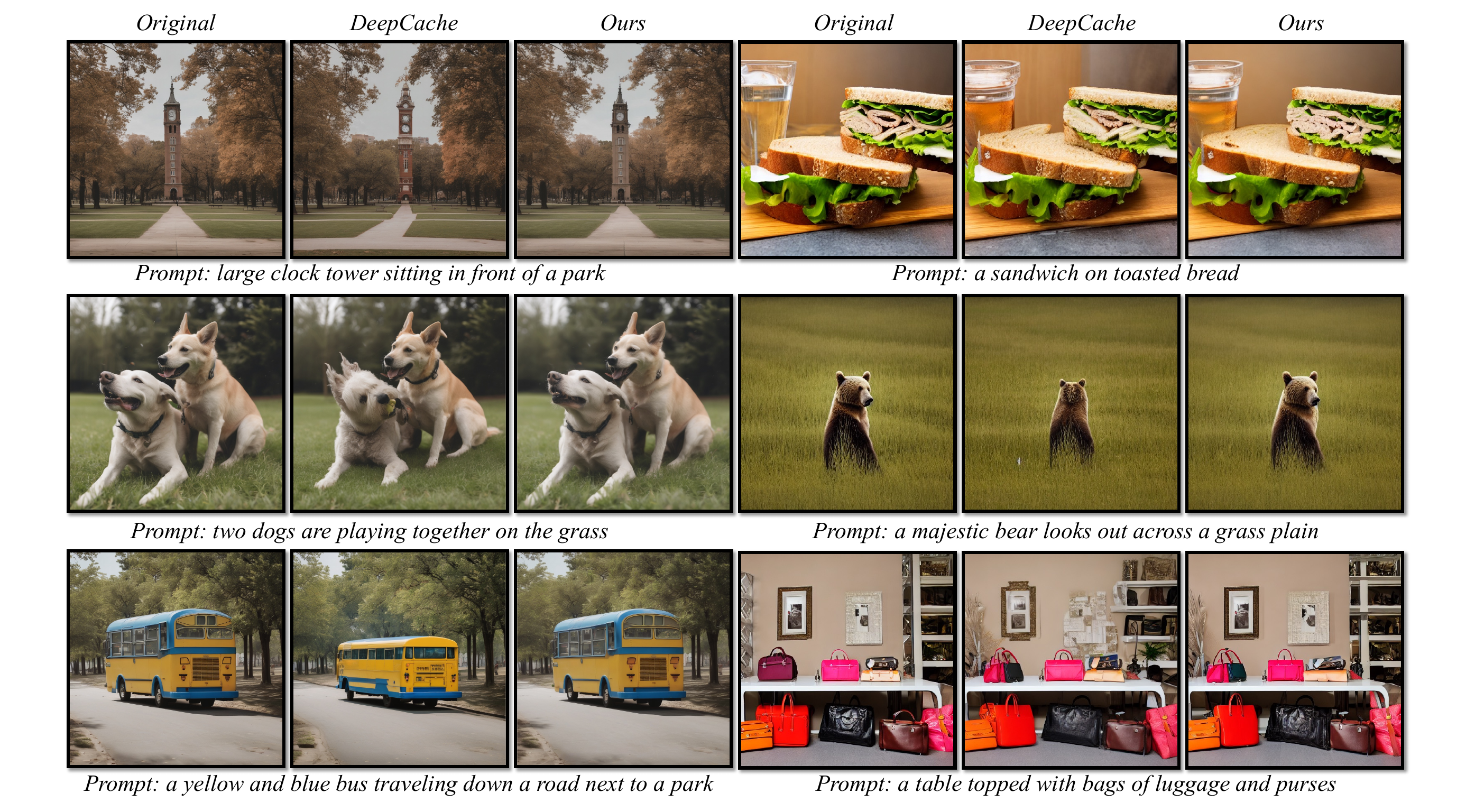}
\vspace{-0.45cm}
\caption{Qualitative results of text-to-image generation task using SDXL and SD-1-5 on MS-COCO 2017 benchmark. Left: SDXL, Right: SD-1-5.}
\label{fig:vis_images}
\vspace{-0.3cm}
\end{figure}

\begin{figure}[tb!]
\centering
\includegraphics[width=1.0\linewidth]{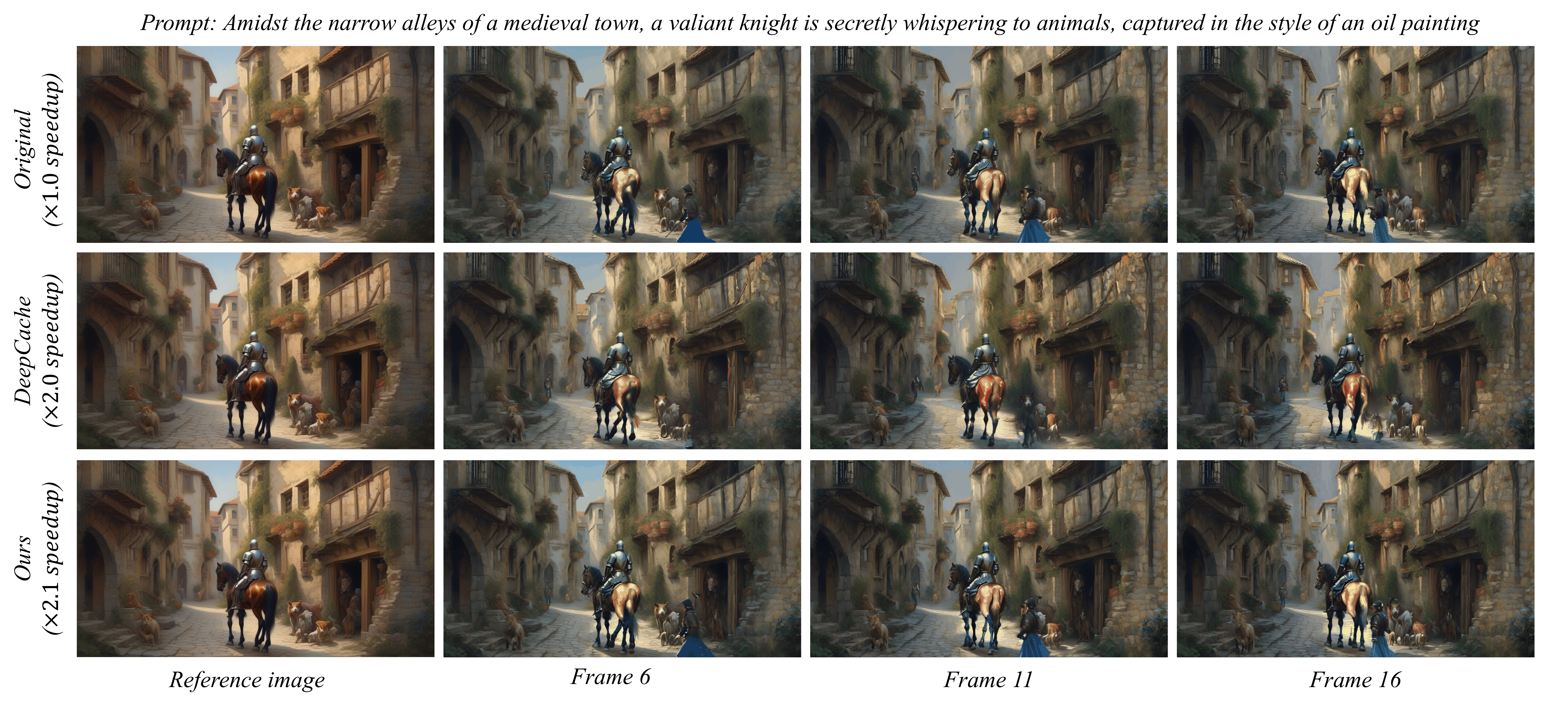}
\vspace{-0.45cm}
\caption{Qualitative results of image-to-video generation task using I2VGen-XL on AIGCBench.}
\label{fig:vis_videos}
\vspace{-0.15cm}
\end{figure}

\begin{figure}[tb!]
\vspace{-0.45cm}
\small
  \centering
  \resizebox{\linewidth}{!}{
  \begin{subfigure}[b]{0.45\textwidth}
    \includegraphics[width=\textwidth]{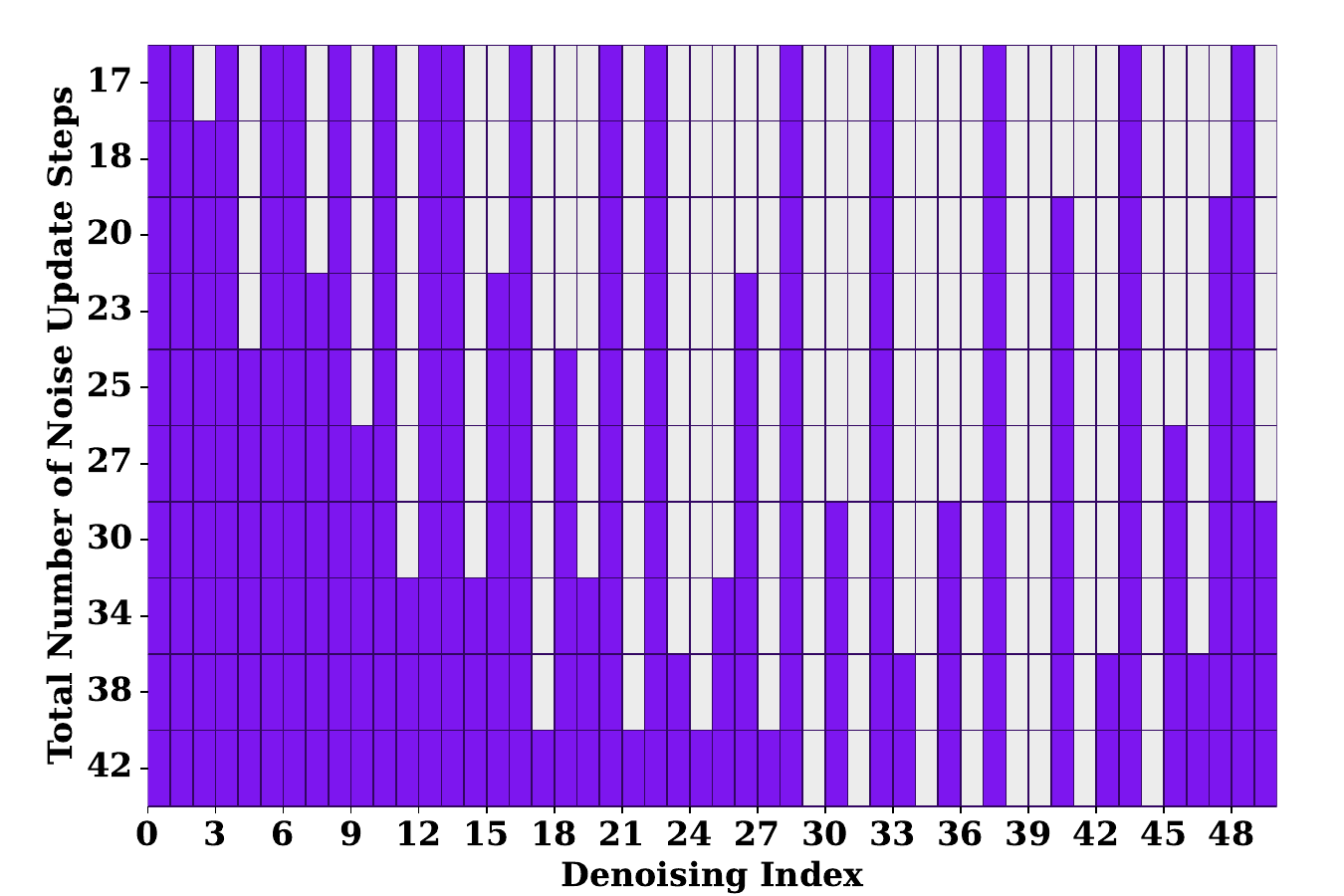}
    \vspace{-1.5em}
    \caption{}
    \label{fig:path_greedy}
  \end{subfigure}
  \begin{subfigure}[b]{0.45\textwidth}
    \includegraphics[width=\textwidth]{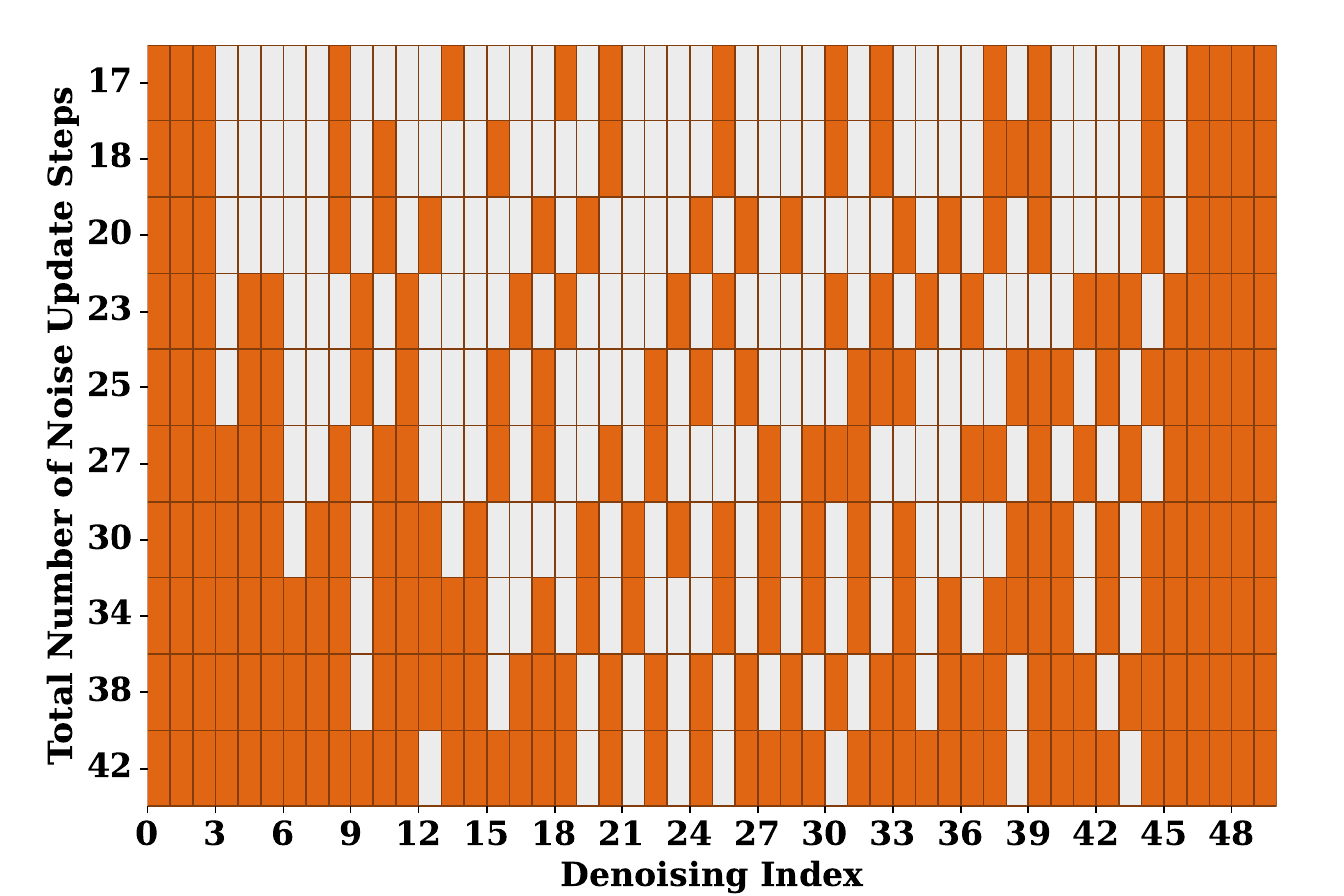}
    \vspace{-1.5em}
    \caption{}
    \label{fig:path_our}
  \end{subfigure}
  \begin{subfigure}[b]{0.45\textwidth}
    \includegraphics[width=\textwidth]{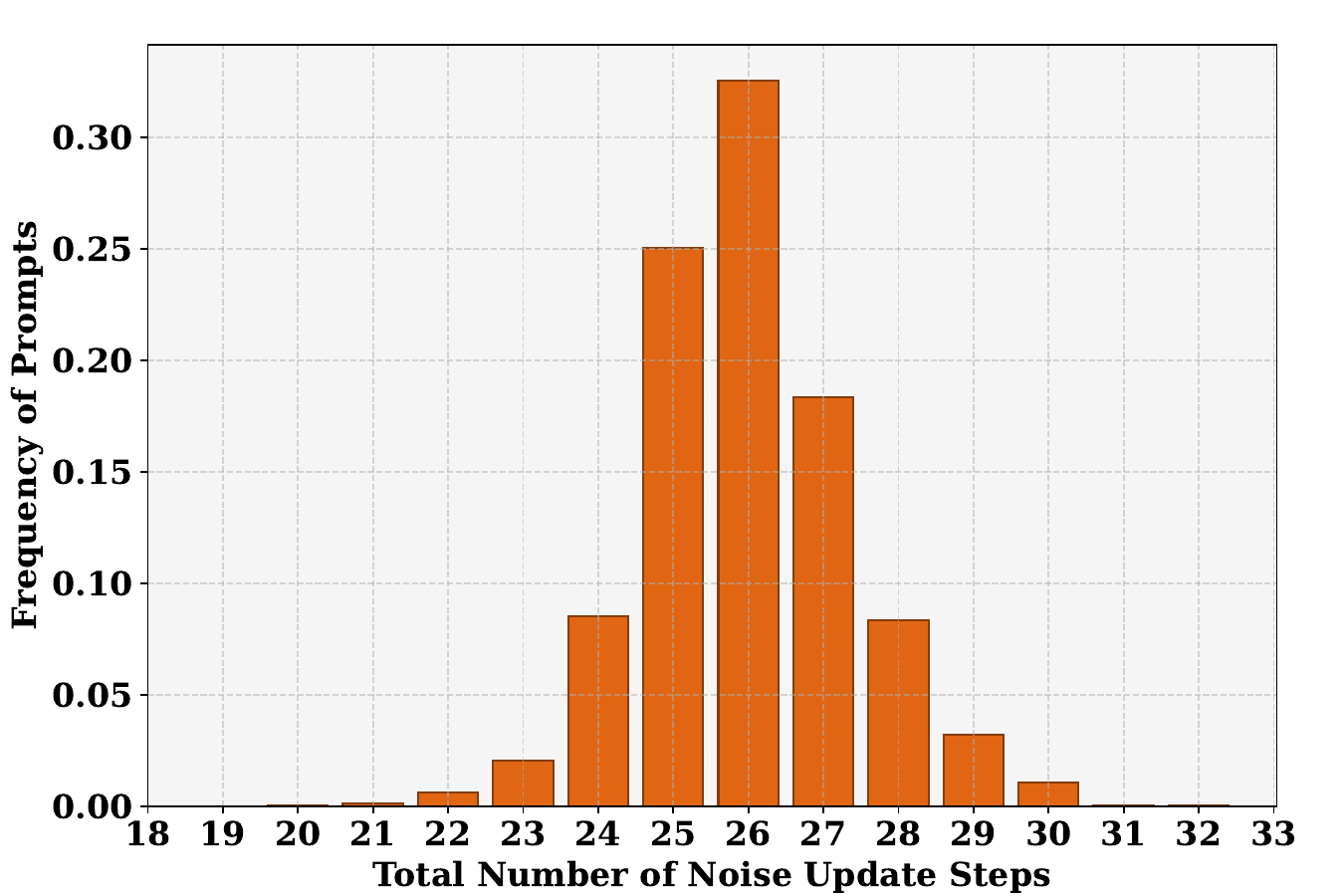}
    \vspace{-1.5em}
    \caption{}
    \label{fig:path_coco}
  \end{subfigure}}
  \vspace{-0.65cm}
  \caption{\small (a) Skipping paths under different skipping targets obtained by the greedy search algorithm. (b) Skipping paths under different skipping thresholds by the third-order estimator. (c) The frequency distribution of the skipping number of noise update steps for SDXL generating images on MS-COCO 2017 benchmark.}
  \label{fig:path_vis}
\vspace{-0.45cm}
\end{figure}

\paragraph{Generation Comparisons.} To demonstrate the effectiveness of AdaptiveDiffusion more intuitively, we show some visualization results. Fig.~\ref{fig:vis_images} shows the results of the text-to-image generation task, it can be seen that AdaptiveDiffuion can better maintain image quality compared to Deepcache with nearly equal acceleration. Since AdaptiveDiffusion can adaptively determine which steps can be skipped, unimportant steps that have little impact on the final generation quality will be skipped during inference. Besides, to demonstrate the generalization of AdaptiveDiffusion on different tasks, we provide video generation results in Fig.~\ref{fig:vis_videos}. More generalization results can be found in Appendix \ref{app:vis}

\vspace{-0.15cm}
\paragraph{Denoising Path Comparisons.} Fig.~\ref{fig:path_vis} illustrates the distribution of skipping paths at different skipping schemes. It can be observed from Fig. \ref{fig:path_greedy} and \ref{fig:path_our} that when the number of noise update steps keeps decreasing (more blank grids in the horizontal lines), both greedy searched paths and third-order estimated paths tend to prioritize the importance of early and late denoising steps. From Fig. \ref{fig:path_coco}, we find that most prompts in the MS-COCO 2017 benchmark only need around 26 steps of noise update to generate an almost lossless image against the 50-step generation result.

\vspace{-0.15cm}
\section{Conclusion}
\vspace{-0.25cm}

In this paper, we explore the training-free diffusion acceleration and introduce AdaptiveDiffusion, which can dynamically select the denoising path according to given prompts. Besides, we perform the error analyses of the step-skipping strategy and propose to use the third-order estimator to indicate the computation redundancy. Experiments are conducted on MS-COCO 2017, ImageNet, AIGCBench and MSR-VTT, showing a good trade-off between high image quality and low inference cost.

\section{Acknowledgements}
The research was supported by the National Key R\&D Program of China (Grant No. 2022ZD0160104), the Science and Technology Commission of Shanghai Municipality (Grant No. 22DZ1100102) Shanghai Rising Star Program (Grant No. 23QD1401000), National Key Research and Development Program of China (No. 2022ZD0160101), Shanghai Natural Science Foundation (No. 23ZR1402900), Shanghai Municipal Science and Technology Major Project (No.2021SHZDZX0103), National Natural Science Foundation of China (No. 62071127, and 62101137).

{
    \small
    \bibliographystyle{plain}
    \bibliography{main}
}

\clearpage
\appendix
\section{Appendix}
Due to the nine-page limitation of the manuscript, we provide more details and visualizations from the following aspects:

\begin{itemize}
    \item Sec.~\ref{app:limitation}: Limitations and Broader Impacts.
    \item Sec.~\ref{app:method_explanation}: Method Explanation.
    \begin{itemize}
        \item Sec.~\ref{app:proof}: Theoretical Proof of Single-step Skipping.
        \item Sec.~\ref{app:two-step}: Theoretical Proof of Two-step Skipping.
        \item Sec.~\ref{app:k-step}: Theoretical Proof of $k$-step Skipping.
        \item Sec.~\ref{app:theoretical_third}: Theoretical Relation between the 3rd-order Estimator and Optimal Skipping Strategy.
        \item Sec.~\ref{app:algorithms}: Algorithms.
    \end{itemize}
    \item Sec.~\ref{app:exp_details}: More Experimental Details.
    \begin{itemize}
        \item Sec.~\ref{app:more_val_details}: More Evaluation Details.
        \item Sec.~\ref{app:prompts}: Prompts in AIGCBench.
    \end{itemize}
    \item Sec.~\ref{app:vis}: More Generation Visualizations.
\end{itemize}

\subsection{Limitations and Broader Impacts}
\label{app:limitation}

Currently, AdaptiveDiffusion is only evaluated on image generation and video generation tasks. A long-term vision is to achieve lossless acceleration on any modalities generation such as 3D and speech. Besides, although AdaptiveDiffusion can improve the speed on few-step (\textit{e.g.}, 10 steps) settings, it may not work for extreme few-step generation for the rapid changes of the latent features.

\subsection{Method Explanations}
\label{app:method_explanation}
\subsubsection{Error Estimation Induced by Single-step Skipping}
\label{app:proof}
In Sec. \ref{sec:error}, the error between the latents of the continuous noise update and one-step skipping of noise update is stated to be upper-bounded of the difference between the previous two output latents. Here, we provide the detailed proof of this statement.
\paragraph{Assumptions}
We first make the assumptions that:\\ 
(1) $\epsilon_\theta(x, t)$ is Lipschitz w.r.t to its paramters $x$ and $t$;\\
(2) The 1st-order difference $\varDelta x_{i}=x_i - x_{i+1}$ exists and is continuous for $0 \leq i \leq T-1$. 

\paragraph{Proof}
Take $i$-th step as an example to perform the one-step skipping of noise prediction, we can obtain the following update formulations.
\small
\begin{equation}
\begin{aligned}
x_{i+1} & =f(i+1) \cdot x_{i+2}- g(i+1) \cdot \epsilon_{\theta}(x_{i+2}, t_{i+2});\\
x_{i} & =f(i) \cdot x_{i+1}-g(i) \cdot \epsilon_{\theta}(x_{i+1},t_{i+1});\\
x_{i-1} & =f(i-1) \cdot x_{i}-g(i-1) \cdot \epsilon_{\theta}(x_{i+1}, t_{i+1});\\
x_{i-2} & =f(i-2) \cdot x_{i-1}-g(i-2) \cdot \epsilon_{\theta}(x_{i-1}, t_{i-1});
\end{aligned}
\end{equation}

Here, the $i$-th step of noise prediction is replaced by reusing the $i+1$-th noise prediction. Then, the error caused by skipping the $i$-th noise update, $\varepsilon_{i-1}=\|x_{i-1}-x_{i-1}^{ori}\|$, can be derived as follows.

\begin{equation}
\small
\begin{aligned}
\|\varepsilon_{i-1}\| &= \|g(i-1)\cdot[\epsilon_{\theta}(x_{i+1}, t_{i+1})-\epsilon_{\theta}(x_{i}, t_{i})]\| \\
&= \|g(i-1) \cdot [\epsilon_{\theta}(x_{i+1}, t_{i+1})-\epsilon_{\theta}(x_{i}, t_{i+1})+\epsilon_{\theta}(x_{i}, t_{i+1})-\epsilon_{\theta}(x_{i}, t_{i})]\|\\
&\leq \|g(i-1) \cdot \mathcal{O}(t_{i}-t_{i+1})\|+\|g(i-1) \cdot \mathcal{O}(x_{i}-x_{i+1})\| \\
&=\mathcal{O}(t_{i}-t_{i+1})+\mathcal{O}(x_{i}-x_{i+1}).
\end{aligned}
\end{equation}
The proof utilizes the property that $\|\epsilon_\theta(x_i, t_{i+1}) - \epsilon_\theta(x_{i+1}, t_{i+1})\|$ and $\|\epsilon_\theta(x_i, t_i) - \epsilon_\theta(x_{i}, t_{i+1})\|$ are upper-bounded by $\mathcal{O}(x_{i}-x_{i+1})$ and $\mathcal{O}(t_{i}-t_{i+1})$ respectively according to the Lipschitz continuity.

\subsubsection{Error Estimation Induced by Two-step Skipping}\label{app:two-step}
We further estimate the situation of consecutive two-step skipping of the noise prediction model. Assuming that noise prediction is skipped at the $i$-th and $(i-1)$-th steps, then we have the following formulation.

\begin{equation}
\begin{aligned}
x_{i+1} & =f(i+1) \cdot x_{i+2}- g(i+1) \cdot \epsilon_{\theta}(x_{i+2}, t_{i+2});\\
x_{i} & =f(i) \cdot x_{i+1}-g(i) \cdot \epsilon_{\theta}(x_{i+1},t_{i+1});\\
x_{i-1} & =f(i-1) \cdot x_{i}-g(i-1) \cdot \epsilon_{\theta}(x_{i+1}, t_{i+1});\\
x_{i-2} & =f(i-2) \cdot x_{i-1}-g(i-2) \cdot \epsilon_{\theta}(x_{i+1}, t_{i+1}).
\end{aligned}
\end{equation}

The error, $\varepsilon_{i-2}=\|x_{i-2}-x_{i-2}^{ori}\|$, can be derived as follows.

\begin{equation}
\begin{aligned}
 \varepsilon_{i-2} &= \|f(i-2)\cdot(x_{i-1}-x_{i-1}^{ori})-g(i-2)\cdot[\epsilon_{\theta}(x_{i+1}, t_{i+1})-\epsilon_{\theta}(x_{i-1}^{ori}, t_{i-1})]\| \\
&= \|f(i-2) \cdot g(i-1) \cdot [\epsilon_{\theta}(x_{i}, t_{i})-\epsilon_{\theta}(x_{i+1}, t_{i+1})]\\
 &\quad-g(i-2) \cdot [\epsilon_{\theta}(x_{i+1}, t_{i+1})-\epsilon_{\theta}(x_{i-1}^{ori}, t_{i-1})]\| \\
&= \|h^2(i-1)\cdot [\epsilon_{\theta}(x_{i},t_{i})-\epsilon_{\theta}(x_{i},t_{i+1})+\epsilon_{\theta}(x_{i},t_{i+1})-\epsilon_{\theta}(x_{i-1}^{ori}, t_{i-1})]\\
&\quad -g(i-2)\cdot [\epsilon_{\theta}(x_{i+1},t_{i+1})-\epsilon_{\theta}(x_{i-1}^{ori},t_{i-1})]\|\\
&\leq \|h^2(i-1)\cdot \mathcal{O}(t_{i}-t_{i+1})\|+\|h^2(i-1)\cdot \mathcal{O}(x_{i}-x_{i+1})\|\\
&\quad +\|g(i-2)\cdot \mathcal{O}(x_{i-1}-x_{i-1}^{ori})\|\\
&= \mathcal{O}(t_{i}-t_{i+1})+\mathcal{O}(x_{i}-x_{i+1})+\mathcal{O}(t_{i-1}-t_{i})+\mathcal{O}(x_{i-1}-x_{i}).
\end{aligned}
\end{equation}
From the above derivations, it can be observed that the skipping error is clearly related to and upper-bounded by the accumulation of previous latent differences. Here $h^{2}\left( i-1 \right)$ is defined as $h^{2}\left( i-1 \right) \coloneqq g\left( i-1 \right) f\left( i-2 \right)$. The above conclusion of skipping error's upper bound can be easily extended to any situation where finite-step skipping is used.

\subsubsection{Error Estimation Induced by $k$-step Skipping}\label{app:k-step}
Take $i$-th step as an example to perform the $k$-step ($k\geq 2$) skipping of noise prediction, we can obtain the following update formulations.
\begin{equation}
\small
\begin{aligned}
x_{i} & =f(i) \cdot x_{i+1}-g(i) \cdot \epsilon_{\theta}(x_{i+1},t_{i+1});\\
x_{i-1}&=f\left( i-1 \right) \cdot x_i-g\left( i-1 \right) \cdot \epsilon _{\theta}\left( x_{i+1}, t_{i+1} \right);\\
x_{i-2}&=f\left( i-2 \right) \cdot x_{i-1}-g\left( i-2 \right) \cdot \epsilon _{\theta}\left( x_{i+1}, t_{i+1} \right);\\
&\vdots\\
x_{i-k}&=f\left( i-k \right) \cdot x_{i-k+1}-g\left( i-k \right) \cdot \epsilon _{\theta}\left( x_{i+1}, t_{i+1} \right).\\
\Rightarrow \varepsilon _{i-k}&=\left\| x_{i-k}-x_{i-k}^{ori} \right\|\\
&=\left\| f\left( i-k \right) \left( x_{i-k+1}-x_{i-k+1}^{ori} \right) -g\left( i-k \right) \left[ \epsilon _{\theta}\left( x_{i+1}, t_{i+1} \right) -\epsilon _{\theta}\left( x_{i-k+1}, t_{i-k+1} \right) \right] \right\|\\
&\leqslant f\left( i-k \right) \varepsilon _{i-k+1}+g\left( i-k \right) \left\| \epsilon _{\theta}\left( x_{i+1}, t_{i+1} \right) -\epsilon _{\theta}\left( x_{i-k+1}, t_{i-k+1} \right) \right\|\\
&\leqslant \sum_{m=1}^{k-1}{\left\| h^{k-m+1}\left( i-m \right) \cdot \mathcal{O} \left( t_{i-m+1}-t_{i-m+2} \right) \right\|}\\
&\quad+\sum_{m=1}^{k-1}{\left\| h^{k-m+1}\left( i-m \right) \cdot \mathcal{O} \left( x_{i-m+1}-x_{i-m+2} \right) \right\|}\\
&\quad+\left\| g\left( i-k \right) \cdot \mathcal{O} \left( x_{i-k+1}-x_{i-k+2} \right) \right\| +\left\| g\left( i-k \right) \cdot \mathcal{O} \left( t_{i-k+1}-t_{i-k+2} \right) \right\|\\
&=\sum_{m=1}^k{\mathcal{O} \left( t_{i-m+1}-t_{i-m+2} \right) +\mathcal{O} \left( x_{i-m+1}-x_{i-m+2} \right)}.\\
\end{aligned}
\end{equation}
The derivation also utilizes the property that $\|\epsilon_\theta(x_i, t_{i+1}) - \epsilon_\theta(x_{i+1}, t_{i+1})\|$ and $\|\epsilon_\theta(x_i, t_i) - \epsilon_\theta(x_{i}, t_{i+1})\|$ are upper-bounded by $\mathcal{O}(x_{i}-x_{i+1})$ and $\mathcal{O}(t_{i}-t_{i+1})$ respectively according to the Lipschitz continuity. Here $h^{k-m+1}\left( i-m \right)$ is defined as $h^{k-m+1}\left( i-m \right) \coloneqq g\left( i-m \right) \prod\nolimits_{j=1}^{k-m}{f\left( i-m-j \right)}$.

From the above derivation, it can be observed that the error of an arbitrary $k$-step skipping scheme is related to and upper-bounded by the accumulation of previous latent differences. Therefore, if the skipping step of noise prediction is large, the upper bound of the error will naturally increase, which is also empirically demonstrated by Fig. \ref{fig:accum_error}.

\subsubsection{Theoretical Relation between the 3rd-order Estimator and Optimal Skipping Strategy}\label{app:theoretical_third}
To explore the theoretical relationship between the third-order estimator and the skipping strategy, we need to formulate the difference between the neighboring noise predictions. According to Eq. (\ref{eq:general_update}), we can get the following first-order differential equations regarding the latent $x$.
\begin{equation}
\small
\begin{aligned}
\varDelta x_i&= x_i-x_{i+1}=\left[ 1-f\left( i \right) \right] x_{i+1}-g\left( i \right) \cdot \epsilon _{\theta}\left( x_{i+1}, t_{i+1} \right);
\\
\varDelta x_{i-1}&= x_{i-1}-x_i=\left[ 1-f\left( i-1 \right) \right] x_i-g\left( i-1 \right) \cdot \epsilon _{\theta}\left( x_i, t_i \right).
\\
\end{aligned}
\end{equation}
Now, let $u\left( i \right) \coloneqq 1-f\left( i-1 \right)$, and we further derive the second-order differential equations based on the above equations.
\begin{equation}
\small
\begin{aligned}
&\varDelta x_{i-1}-\varDelta x_i\\
=&u\left( i \right) x_i-u\left( i+1 \right) x_{i+1}+g\left( i \right) \cdot \epsilon _{\theta}\left( x_{i+1}, t_{i+1} \right) -g\left( i-1 \right) \cdot \epsilon _{\theta}\left( x_i, t_i \right) 
\\
=&u\left( i \right) \left( x_i-x_{i-1} \right) +u\left( i \right) x_{i-1}-u\left( i+1 \right) \left( x_{i+1}-x_i \right) -u\left( i+1 \right) x_i+g\left( i \right) \cdot \epsilon _{\theta}\left( x_{i+1}, t_{i+1} \right)
\\
&-g\left( i-1 \right) \cdot \epsilon _{\theta}\left( x_i, t_i \right) 
\\
=&u\left( i \right) \varDelta x_{i-1}-u\left( i+1 \right) \varDelta x_i+\varDelta \left[ u\left( i \right) x_{i-1} \right] +g\left( i \right) \cdot \epsilon _{\theta}\left( x_{i+1}, t_{i+1} \right) -g\left( i-1 \right) \cdot \epsilon _{\theta}\left( x_i, t_i \right) 
\\
=&u\left( i \right) \varDelta x_{i-1}-u\left( i+1 \right) \varDelta x_i+\varDelta \left[ u\left( i \right) x_{i-1} \right] +g\left( i \right) \left[ \epsilon _{\theta}\left( x_{i+1}, t_{i+1} \right) -\epsilon _{\theta}\left( x_i, t_i \right) \right] 
\\
&+\left[ g\left( i \right) -g\left( i-1 \right) \right] \epsilon _{\theta}\left( x_i, t_i \right) 
\\
=&u\left( i \right) \varDelta x_{i-1}-u\left( i+1 \right) \varDelta x_i+\varDelta \left[ u\left( i \right) x_{i-1} \right] -g\left( i \right) \varDelta \epsilon _{\theta}^{i}-\varDelta g\left( i \right) \cdot \epsilon _{\theta}\left( x_i, t_i \right).
\end{aligned}
\end{equation}
After simplification of the above equation, we can get the following formulation:
\begin{equation}
\small
f\left( i-1 \right) \varDelta x_{i-1}-f\left( i \right) \varDelta x_i=\varDelta \left[ u\left( i \right) x_{i-1} \right] -g\left( i \right) \varDelta \epsilon _{\theta}^{i}-\varDelta g\left( i \right) \cdot \epsilon _{\theta}\left( x_i, t_i \right).
\end{equation}
From the above equation, we can observe that the difference between noise predictions $\varDelta \epsilon _{\theta}^{i}$ is related to the first- and second-order derivatives of $x_i$, as well as the noise prediction $\epsilon _{\theta}\left( x_i, t_i \right)$. Therefore, it would be difficult to estimate the difference without $\epsilon _{\theta}\left( x_i, t_i \right)$. Now we consider the third-order differential equation. From the above equation, we further obtain the following formulation.
\begin{equation}
\small
\begin{aligned}
&f\left( i \right) \varDelta x_i-f\left( i+1 \right) \varDelta x_{i+1}=\varDelta \left[ u\left( i+1 \right) x_i \right] -g\left( i+1 \right) \varDelta \epsilon _{\theta}^{i+1}-\varDelta g\left( i+1 \right) \cdot \epsilon _{\theta}\left( x_{i+1}, t_{i+1} \right);\\
\Rightarrow &\varDelta \left[ f\left( i-1 \right) \varDelta x_{i-1} \right] -\varDelta \left[ f\left( i \right) \varDelta x_i \right] =\varDelta ^{\left( 2 \right)}\left[ u\left( i \right) x_{i-1} \right] -\varDelta \left[ g\left( i \right) \varDelta \epsilon _{\theta}^{i} \right] -\varDelta \left[ \varDelta g\left( i \right) \cdot \epsilon _{\theta}\left( x_i, t_i \right) \right];\\
\Rightarrow &\varDelta \left[ \varDelta g\left( i \right) \cdot \epsilon _{\theta}\left( x_i, t_i \right) \right] =-\varDelta ^{\left( 2 \right)}\left[ f\left( i-1 \right) \varDelta x_{i-1} \right] +\varDelta ^{\left( 2 \right)}\left[ u\left( i \right) x_{i-1} \right] -\varDelta \left[ g\left( i \right) \varDelta \epsilon _{\theta}^{i} \right].\\
\end{aligned}
\end{equation}
From the above equation, it can be observed that the difference of the neighboring noise predictions is explicitly related to the third- and second-order derivatives of $x_i$, as well as the second-order derivative of $\epsilon_\theta^{i}$. Since $\lim_{i \rightarrow 0} f\left( i \right) =1,\lim_{i \rightarrow 0} u\left( i \right) =0,\lim_{i \rightarrow 0} g\left( i \right) =0$, we can finally get the conclusion that $\varDelta \epsilon _{\theta}^{i}|_{i \rightarrow 0} = \mathcal{O} \left( \varDelta ^{\left( 3 \right)}x_{i-1} \right)$.

\subsubsection{Algorithms}
\label{app:algorithms}

\noindent \textbf{Algorithm of greedy search for optimal skipping path.} The specific details of the greedy search algorithm used as the optimal denoising path are shown in Alg. (\ref{alg:greedy}).

\begin{algorithm}[H]
\caption{Greedy Search for the Optimal Skipping Path.}
\label{alg:greedy}
\renewcommand{\algorithmicensure}{\textbf{Input:}}
\begin{algorithmic}[1]
\Ensure{Noise Prediction Model $\epsilon_\theta$, Sampling Scheduler $\phi$, Decoder $\mathcal{F}_d$, Target Skipping Step Number $N$, Sample Step $T$, Conditional embedding $c$;}
\State Initialize Skipping Path $\mathcal{S}$ = [True] * $T$, Current Skipping Step Number $n_{skip}$ = 0.
\State Compute $x_0^{ori}$ by Eq. (\ref{eq:general_update}).
\While {$n_{skip} < N$}
\State \begin{varwidth}[t]{\linewidth}
Initialize $df$ = [];
\end{varwidth}
\For {$i$ in range($T - 1$)}
\If{$\mathcal{S}[i]$==True}
\State \begin{varwidth}[t]{0.95\linewidth}
temp\_path = $\mathcal{S}$.copy();
\end{varwidth}
\State \begin{varwidth}[t]{0.95\linewidth}
temp\_path[$i$]=False;
\end{varwidth}
\State \begin{varwidth}[t]{0.95\linewidth}
Generate $x_0^{temp}$ by Eq. (\ref{eq:skip_strategy});
\end{varwidth}
\State \begin{varwidth}[t]{0.95\linewidth}
Compute $\ell_1$ difference $\mathcal{L}_0 = \|x_0^{temp} - x_0^{ori}\|$;
\end{varwidth}
\State \begin{varwidth}[t]{0.95\linewidth}
$df$.append($\mathcal{L}_0$);
\end{varwidth}
\EndIf
\EndFor
\State \begin{varwidth}[t]{0.95\linewidth}
index = argmin($df$);
\end{varwidth}
\State \begin{varwidth}[t]{0.95\linewidth}
Set $\mathcal{S}[$index$]$ = False;
\end{varwidth}
\EndWhile\\
\Return{$\mathcal{S}$.}
\end{algorithmic}
\end{algorithm}

\noindent \textbf{Algorithm of the overall skipping strategy in AdaptiveDiffusion.} The designed skipping strategy used in our AdaptiveDiffusion are elaborated in Alg. (\ref{alg:adaptive}).

\begin{algorithm}[H]
\caption{AdaptiveDiffusion.}
\label{alg:adaptive}
\renewcommand{\algorithmicensure}{\textbf{Input:}}
\begin{algorithmic}[1]
\Ensure{Noise Prediction Model $\epsilon_\theta$, Sampling Scheduler $\phi$, Decoder $\mathcal{F}_d$, Sample Step $T$, Conditional embedding $c$, Maximum Skipping Step Number $C_{\text{max}}$, Threshold $\delta$;}
\State Initialize Random Noise $x$, Skipping Path $\mathcal{S}$ = [], Previous Differential List $P_{\text{diff}}$ = [], Previous Latent List $P_{\text{latent}}$ = [].
\For {$i$ in range($T - 1$)}
\If{$i\leq 2$}
\State \begin{varwidth}[t]{0.95\linewidth}
$\mathrm{O}_{pv}$ = $\mathrm{O}$ = $\epsilon(x, i)$;
\end{varwidth}
\If{$i\geq 1$}
\State \begin{varwidth}[t]{0.95\linewidth}
$P_{\text{diff}}$.append($\|x-P_{\text{latent}}[-1]\|$);
\end{varwidth}
\EndIf
\Else
\If{$\mathcal{S}[-1]$==True}
\State \begin{varwidth}[t]{0.95\linewidth}
$\mathrm{O}_{pv}$ = $\mathrm{O}$ = $\epsilon(x, i)$;
\end{varwidth}
\Else
\State \begin{varwidth}[t]{0.95\linewidth}
$\mathrm{O}$ = $\mathrm{O}_{pv}$;
\end{varwidth}

\EndIf

\EndIf
\State \begin{varwidth}[t]{0.95\linewidth}
Compute $\phi(x)$ by Eq. (\ref{eq:general_update});
\end{varwidth}
\If{$i\geq 3$}
\State \begin{varwidth}[t]{0.95\linewidth}
$\mathcal{S}$.append($\longleftarrow \xi(x)$) in Eq. (\ref{eq:criterion});
\end{varwidth}
\EndIf
\State \begin{varwidth}[t]{0.95\linewidth}
$P_{\text{latent}}$.append($x$);
\end{varwidth}

\EndFor
\\
\Return{$\mathcal{F}_d(x)$.}
\end{algorithmic}
\end{algorithm}

\subsection{Experimental Details}
\label{app:exp_details}

\subsubsection{More Evaluation Details}
\label{app:more_val_details}

\paragraph{Statistical Analysis of the Estimated Skipping Path.} In Fig. \ref{fig:estimator_effectiveness}, we aim to evaluate the correlation between the estimated skipping paths by our method and the searched optimal skipping paths under different target skipping numbers. This evaluation is crucial to validate the effectiveness of our approach in accurately predicting skipping paths. The paths are represented as sequences of binary choices (0 for a skipping step, 1 for a non-skipping step) at each timestep. We employ a $\chi^2$ test of independence to assess the statistical correlation between the estimated and optimal paths. We first randomly select several prompts from the test set and use the greedy search algorithm to greedily search the optimal skipping paths under different skipping target steps. For the estimated skipping paths, we discard the setting of $C_{\text{max}}$ to test the effectiveness of the third-order estimation without regularization and refine the threshold $\delta$ in a wide range to achieve various skipping paths with different skipping step numbers. Then, for each skipping step number, we construct a 2$\times$2 contingency table from the searched optimal skipping paths and the estimated skipping paths, with the same skipping step numbers. Finally, using the contingency table, we compute the $\chi^2$ statistics and the corresponding $p$-value to assess the independence between the estimated and the searched optimal skipping paths.

\paragraph{Evaluation Tools and Details.} Following~\citep{li2024distrifusion}, we use TorchMetrics~\footnote{\url{https://github.com/Lightning-AI/torchmetrics}} to calculate PSNR and LPIPS, and use CleanFID~\citep{parmar2021cleanfid}\footnote{\url{https://github.com/GaParmar/clean-fid}} to calculate FID. Since our method is an adaptive method which means it can choose a suitable skipping path for different prompts, we set batch size to 1 during evaluation. The MACs reported in our experiments are the total MACs in the denoising process following~\citep{li2024distrifusion}. Besides, the reported MACs, speedup ratio, and latency are the average over the whole dataset. 

\paragraph{CodeBase.} For a fair comparison, when compared with Adaptive DPM-Solver, we use the official codebase of DPM-Solver\footnote{\url{https://github.com/LuChengTHU/dpm-solver}}. Experiments for LDM-4-G are based on the official LDM codebase\footnote{\url{https://github.com/CompVis/latent-diffusion}} and DeepCache codebase\footnote{\url{https://github.com/horseee/DeepCache}}. Other experiments are based on Diffusers\footnote{\url{https://github.com/huggingface/diffusers}}.

\paragraph{Details of Reproduced DeepCache.} In this paper, we mainly compare our AdaptiveDiffusion with DeepCache. To ensure the generation quality, we set \verb|cache_interval=3| and \verb|cache_branch_id=3| for SD-v1-5, SDXL, and ModelScope. Since the memory cost will be largely improved on video generation tasks especially high-resolution tasks (\textit{e.g.}, I2VGen-XL), We set \verb|cache_interval=3| and \verb|cache_branch_id=0| to reduce the memory costs. For class-conditional image generation task, we follow the official setting and set \verb|cache_interval=10|.

\subsubsection{Prompts in AIGCBench}
\label{app:prompts}
\begin{figure}[t]
\centering
\includegraphics[width=1.0\linewidth]{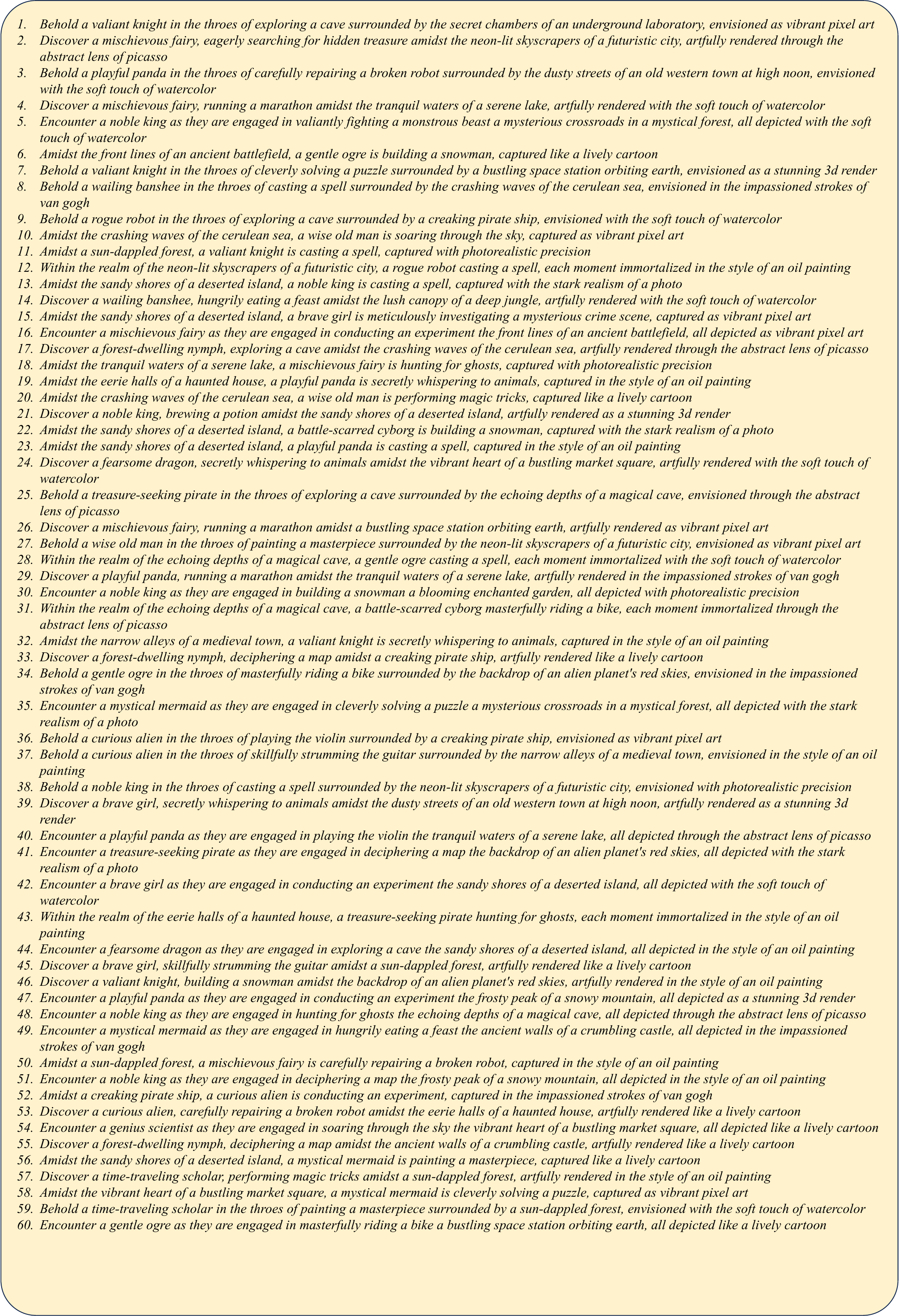}
\caption{Prompts used in AIGCBench.}
\label{fig:prompt_1}
\end{figure}

\begin{figure}[t]
\centering
\includegraphics[width=1.0\linewidth]{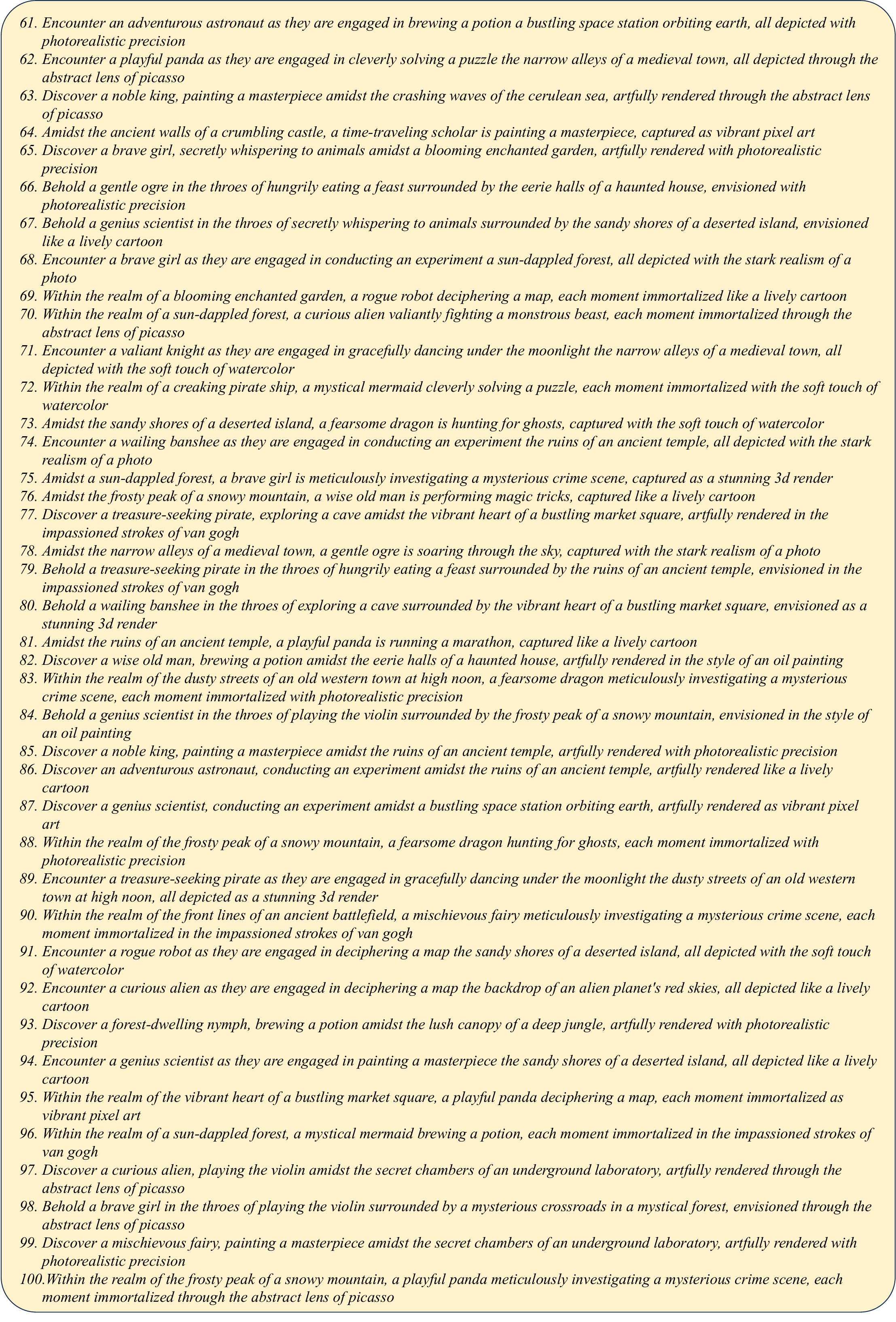}
\caption{Prompts used in AIGCBench.}
\label{fig:prompt_2}
\end{figure}

We randomly sample 100 prompts and their corresponding images as the test set for I2V task. Here, we give the list of the selected prompts in AIGCBench in Fig.~\ref{fig:prompt_1} and Fig.~\ref{fig:prompt_2}. The selected test set includes different styles of images and prompts such as animation, realism, and oil painting.

\subsubsection{Experiments on Unconditional Image Generation}
We provide the acceleration performance of different methods on pure image generation for further comparisons. Specifically, following the experimental setting in Deepcache, we conduct unconditional image generation experiments on CIFAR10 \citep{cifar} and LSUN \citep{yu15lsun} datasets. As shown in Table \ref{tab:unconditional_image}, our method achieves a larger speedup ratio and higher image quality than Deepcache on both benchmarks.

\begin{table}[ht]
\small
\centering
\caption{Performance on Unconditional Image Generation.}
\begin{tabular}{cccc}
\toprule
Datasets                 & Methods               & FID $\downarrow$      & Speedup Ratio\\ \midrule
\multirow{2}{*}{CIFAR10} & Deepcache             & 10.17                 & 2.07x  \\ 
                         & Ours                  & \textbf{7.97}                  & \textbf{2.09x}   \\ \midrule
\multirow{2}{*}{LSUN}    & Deepcache             & 9.13                  & 1.48x  \\ 
                         & Ours                  & \textbf{7.96}                  & \textbf{2.35x} \\ \bottomrule
\end{tabular}
\label{tab:unconditional_image}
\end{table}

\subsubsection{Effectiveness on SDE Solver}
We also explore the effectiveness of our work on SDE solver. Compared with the ODE solver, the SDE solver includes an additional noise item for the latent update, which is unpredictable by previous randomly generated noises. When the magnitude of random noise is not ignorable, the third-order derivative of the neighboring latents cannot accurately evaluate the difference between the neighboring noise predictions. Therefore, to apply our method to SDE solvers, we should design an additional indicator that decides whether the randomly generated noise is minor enough or relatively unchanged to trigger the third-order estimator. In this case, we design an additional third-order estimator for the scaled randomly generated noise. When the third-order derivatives of both the latent and the scaled randomly generated noises are under the respective threshold, the noise prediction can be skipped by reusing the cached noise.

To validate the effectiveness of our improved method, we conduct experiments for SDXL with the SDE-DPM solver on COCO2017. The results are shown in the following table. Compared with Deepcache, our method can achieve higher image quality with a comparable speedup ratio, indicating the effectiveness of AdaptiveDiffusion on SDE solvers.

\begin{table}[ht]
\centering
\small
\caption{Performance on SDE-DPM Solver.}
\begin{tabular}{cccccc}
\toprule
Method    & PSNR  & LPIPS & FID  & Latency (s) & Speedup Ratio \\ \midrule
Deepcache & 16.44 & 0.346 & 8.15 &  9.2    &   1.63x       \\
Ours      & \textbf{18.80} & \textbf{0.232} & \textbf{6.03} &  9.8    &   1.53x         \\ \bottomrule
\end{tabular}
\end{table}

\subsection{More Generation Visualizations}
\label{app:vis}

\begin{figure}[h]
\centering
\includegraphics[width=1.0\linewidth]{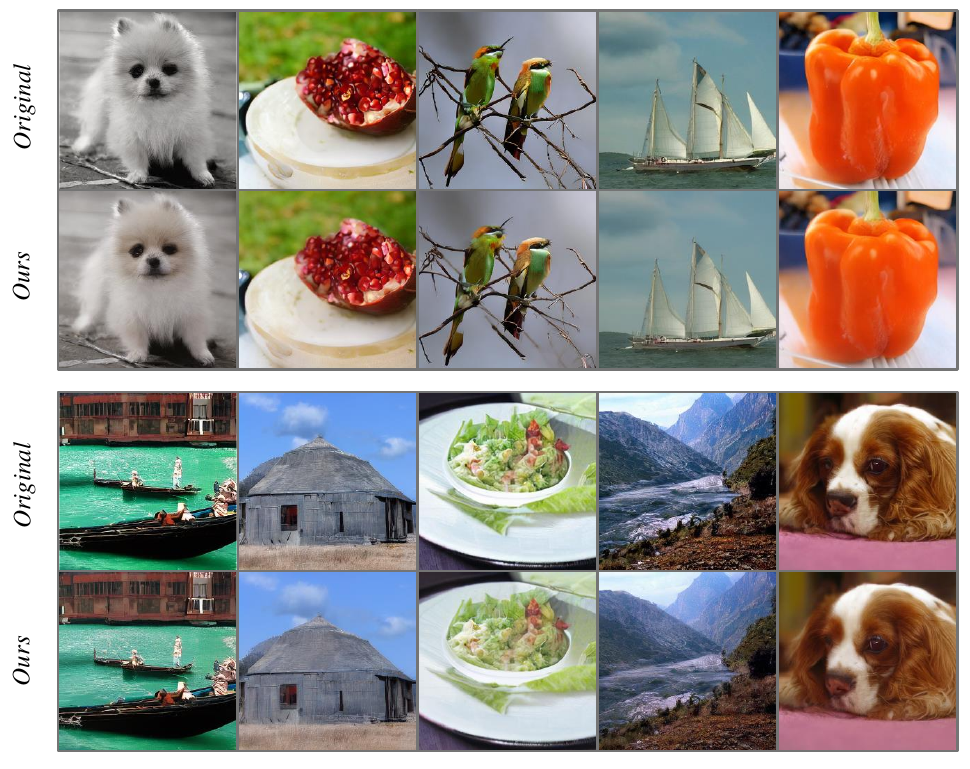}
\vspace{-0.45cm}
\caption{Qualitative results of text-to-image generation task using LDM-4 on ImageNet 256x256 benchmark.}
\label{fig:ldm_vis}
\end{figure}

\begin{figure}[h]
\centering
\includegraphics[width=1.0\linewidth]{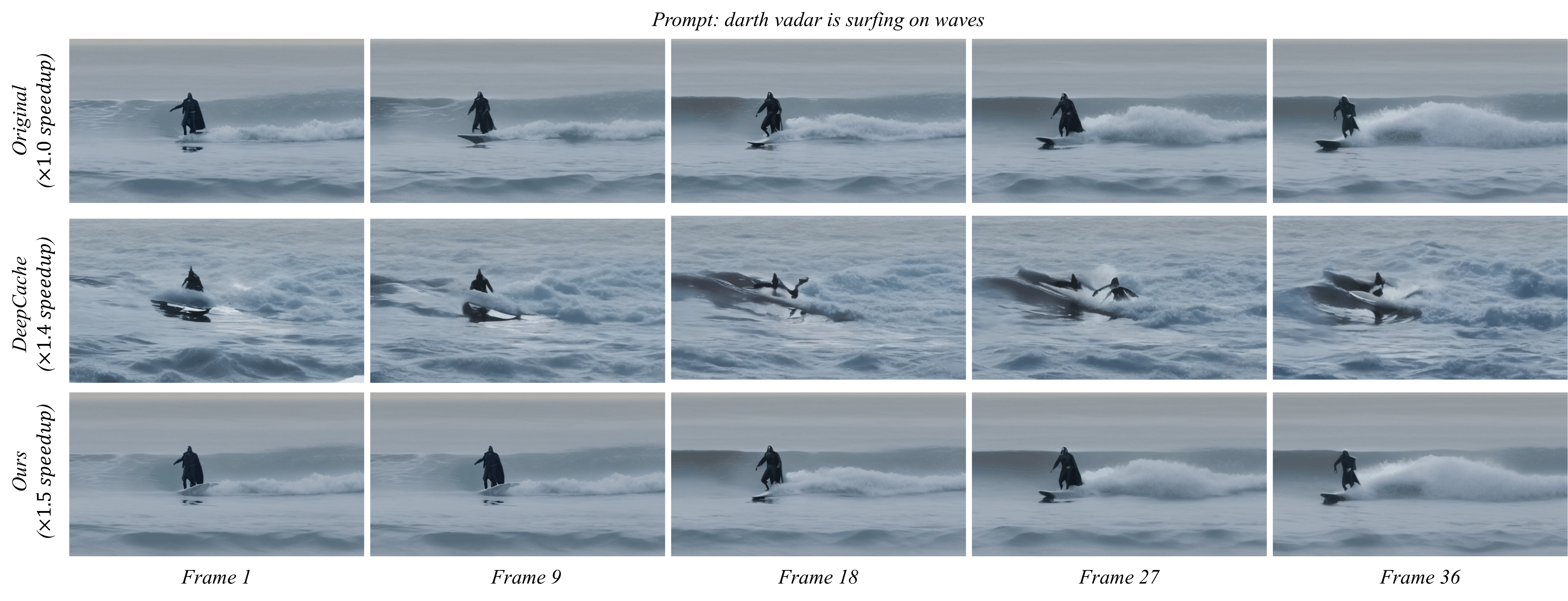}

\vspace{-0.25cm}
\caption{Qualitative results of text-to-video generation task using ModelScopeT2V on MSR-VIT benchmark.}
\label{fig:vis_videos_2}
\end{figure}

Here, we further show more qualitative results tested on text-to-image task using LDM-4 on MS-COCO 2017 and text-to-video generation task using ModelScopeT2V on MSR-VIT, which are illustrated in Fig.~\ref{fig:ldm_vis} and Fig.~\ref{fig:vis_videos_2} respectively.

\end{document}